\documentclass[sigplan,10pt]{acmart}

\acmSubmissionID{<123>}
\usepackage{tikz}
\usepackage{amsmath}
\usepackage{amsthm}

\usepackage{filecontents}

\usepackage[english]{babel}
\usepackage{blindtext}
\usepackage{xcolor}
\usepackage{gincltex}

\usepackage{pgfplots, siunitx, tikzscale}
\usepackage{graphicx}
\usepackage{subfigure}
\usepackage{pifont}
\usetikzlibrary{pgfplots.statistics}
\usepackage{float}
\usepackage{tcolorbox}
\usepackage { bb m}
\usepackage{enumitem}
\usepackage{booktabs}
\usepackage{makecell}
\usepackage {pifont}

\usepackage{amssymb}
\usepackage{algorithm}
\usepackage{algpseudocode}
\usepackage{tabularx}
\usepackage{multirow}
\usepackage{multicol}
\usepackage[
textfont=it
]{caption}

\newcommand{\name}{VaLoRA\xspace}
\newcommand{\hide}[1] {}

\definecolor{mypink1}{rgb}{0.858, 0.188, 0.478}

\newcommand{\tb}{\textcolor{black}}
\newcommand{\tr}{\textcolor{black}}

\vspace{0.4em}
\newcommand{\ie}{\emph{i.e.,}\xspace}
\newcommand{\eg}{\emph{e.g.,}\xspace}
\newcommand{\vs}{\emph{v.s.}\xspace}

\usepackage{titlesec}
\titleformat{\subsection}
  {\normalfont\fontsize{10.5}{12}\bfseries}{\thesubsection}{1em}{}
\titleformat{\subsubsection}[runin]{\normalfont\bfseries}{\thesubsubsection}{1em}{}
\titlespacing*{\subsubsection}{0pt}{\dimexpr 1ex minus 0.7ex}{\dimexpr 1ex plus 0.7ex}

\begin{document}


\title{Empower Vision Applications with LoRA LMM}

\author{
Liang Mi$^1$\footnotemark[1]\footnotemark[2]
Weijun Wang$^2$\footnotemark[1]
Wenming Tu$^2$\footnotemark[2]
Qingfeng He$^2$\footnotemark[2]
Rui Kong$^2$\footnotemark[2]
Xinyu Fang$^2$\footnotemark[2]
Yazhu Dong$^2$\footnotemark[2]
Yikang Zhang$^1$
Yuanchun Li$^{2,3,4}$
Meng Li$^1$
Haipeng Dai$^1$
Guihai Chen$^1$
Yunxin Liu$^{2,3}$\\
$^1$State Key Laboratory for Novel Software Technology, Nanjing University\\
$^2$Institute for AI Industry Research (AIR), Tsinghua University\\
$^3$Shanghai AI Laboratory
$^4$Beijing Academy of Artificial Intelligence (BAAI)
 } 
\renewcommand{\shortauthors}{Mi et al.}

\setcopyright{rightsretained}
\acmYear{2025}\copyrightyear{2025}
\acmConference[EuroSys '25]{Twentieth European Conference on Computer Systems}{March 30--April 3, 2025}{Rotterdam, Netherlands}
\acmPrice{}
\acmBooktitle{Twentieth European Conference on Computer Systems (EuroSys '25), March 30--April 3, 2025, Rotterdam, Netherlands}
\acmDOI{10.1145/3689031.3717472}
\acmISBN{979-8-4007-1196-1/2025/03}

\begin{abstract}
Large Multimodal Models (LMMs) have shown significant progress in various complex vision tasks with the solid linguistic and reasoning capacity inherited from large language models (LMMs).
Low-rank adaptation (LoRA) offers a promising method to integrate external knowledge into LMMs, compensating for their limitations on domain-specific tasks.
However, the existing LoRA model serving is excessively computationally expensive and causes extremely high latency.
In this paper, we present an end-to-end solution that empowers diverse vision tasks and enriches vision applications with LoRA LMMs.
Our system, \name, enables accurate and efficient vision tasks by 1) an accuracy-aware LoRA adapter generation approach that generates LoRA adapters rich in domain-specific knowledge to meet application-specific accuracy requirements, 
2) an adaptive-tiling LoRA adapters batching operator that efficiently computes concurrent heterogeneous LoRA adapters, 
and 3) a flexible LoRA adapter orchestration mechanism that manages application requests and LoRA adapters to achieve the lowest average response latency.
We prototype \name on \tr{five} popular vision tasks on three LMMs.
\tr{Experiment results reveal that \name improves 24-62\% of the accuracy compared to the original LMMs and reduces 20-89\% of the latency compared to the state-of-the-art LoRA model serving systems.}
\end{abstract}


\maketitle

\renewcommand{\thefootnote}{\fnsymbol{footnote}} 
\footnotetext[1]{Liang Mi and Weijun Wang contributed equally to this work.} 
\footnotetext[2]{Work was done while Liang, Wenming, Qingfeng, Rui, Xinyu, and Yazhu interned at the Institute for AI Industry
Research (AIR), Tsinghua University.} 
\footnotetext{
Corresponding authors: Weijun Wang and Meng Li.
}

\section{Introduction}

Encouraged by the success of
LLMs in NLP applications~\cite{LLM12App, 7AppLLM, indataAppLLM, LLMUseCase, li2024personal},  
Large Multimodal Models (LMMs)~\cite{zhu2023minigpt, gpt4o, claude} 
have attracted great attention from both academia and industry.
They enhance LLMs by perceiving and interpreting multimodal signals (\eg visual inputs~\cite{liu2024visual, bai2023qwen, team2023gemini}) and well accomplish many complex multi-modal tasks that prior models cannot.
For example, GPT-4o~\cite{gpt4o} achieves leading accuracy on many multimodal tasks such as 
visual question answering~\cite{goyal2017making}.  
Yet when applied to practical applications requiring domain-specific knowledge, LMMs often show suboptimal performance, similar to the early LLMs that experienced hallucinations~\cite{zhang2023siren}.
\textit{Low-rank adaptation (LoRA)}~\cite{hu2021lora, dettmers2024qlora} provides a promising way to integrate the external knowledge into LMM. 
It fine-tunes a small portion of model parameters, known as LoRA adapters, on domain-specific datasets to learn target knowledge, and freezes the base model to preserve its original capability (more in \S \ref{sec:backgound}).
LLMs often leverage retrieval-augmented generation (RAG)~\cite{lewis2020retrieval} to meet this goal.
Unlike LoRA, which modifies model parameters, RAG appends the retrieved knowledge (\eg documents) onto requests (\ie input data) for accurate response. 
However, this data-augmented method is not appropriate for time-sensitive vision applications. 
Its retrieval process and appended long-context requests incur >10$\times$ response delay~\cite{jin2024ragcache}.
Conversely, LoRA merges fine-tuned adapters into LMMs at runtime and efficiently generates high-quality and consistent domain-specific responses without additional overhead.








Despite these advantages of LoRA, it introduces complex system challenges.
Recent work~\cite{chen2024punica, sheng2023slora, wu2024dlora}, focusing on system optimization for linguistic applications with LoRA LLM, has made noticeable progress.
Punica~\cite{chen2024punica} and S-LoRA~\cite{sheng2023slora} propose \emph{unmerged inference} to overcome the limitation of \emph{merged inference} that can merge only one adapter at once.
It computes multiple LoRA adapters in parallel while batching the shared base model computation across different requests, to boost system efficiency.   
dLoRA~\cite{wu2024dlora} further balances the throughput and latency by merged and unmerged inference mode switch (more in \S \ref{sec:backgound}).
However, these efforts fail to meet the high efficiency and diverse requirements of vision applications (more in \S \ref{sec:challenges}).


In this paper, we answer the following research question:
\textit{Can we leverage LoRA LMMs to enrich vision applications while meeting their performance requirements?}
We argue that yes, but need to tackle the following challenges.

First, many existing small models, trained on domain-specific datasets and used in current vision applications, outperform LMMs on target tasks.
To keep accuracy, their knowledge must be integrated into LMM with LoRA adapters.
However, simply training one LoRA adapter for each task is uneconomical, while fusing too much external knowledge (\eg multiple small models) into a single adapter causes inevitable accuracy degradation.

Second, a vision application often involves diverse external knowledge. 
When serving multiple vision applications, in particular, LoRA LMM is very likely asked to execute multiple LoRA adapters simultaneously.
Although unmerged inference can compute heterogeneous adapters in parallel, they introduce excessive delays. 
Therefore, our LoRA LMM inference system must efficiently compute concurrent heterogeneous adapters with low latency.

Lastly, different vision applications have distinct performance requirements. 
Real-time video analytics application \cite{yuan2022PacketGame, AccDecoder} needs low latency, while visual retrieval~\cite{kang13blazeit} prefers high throughput.
To meet these needs, carefully managing LoRA adapters and flexibly scheduling application requests is necessary.
Typical LoRA adapter manager (\eg dLoRA) is not designed for vision applications, which incurs excessive overhead. 
Solely its inference mode switcher costs over half of LMM inference time.

This paper presents \name, an end-to-end LoRA LMM serving system, including LoRA adapter preparation (\S\ref{sec:LoRAGeneration}) and inference runtime (\S\ref{sec:LoRABatching}, \S\ref{sec:LoRAOrchestration}), to empower vision applications. 
\name addresses the above challenges with the following techniques:



\textit{Accuracy-aware LoRA adapter generation.}
We propose a LoRA generator (\S \ref{sec:LoRAGeneration}) that prepares domain-specific LoRA adapters to generate accurate results on target tasks with external knowledge.
Considering the complex accuracy variations of knowledge fusion (\S \ref{sec:challenges}),  LoRA adapters generation can be formulated as a constrained bin packing problem, that given external knowledge, \ie small models and domain-specific datasets, to generate the minimum number of LoRA adapters, 
ensuring the accuracy specified by vision applications.
We design an accuracy-aware knowledge-fusion algorithm with a greedy heuristic to solve it.
Additionally, we introduce vision task heads, incorporated as part of the LoRA adapter, enabling low-latency response for vision tasks.



\textit{Adaptive-tiling LoRA adapters batching.}
We propose a concurrent LoRA adapters batching method (\S~\ref{sec:LoRABatching}), comprised of the Adaptive-Tiling Matrix Multiplication (ATMM) operator and its optimal tiling search algorithm, for efficient heterogeneous LoRA adapters computation.
The offline search algorithm identifies the optimal tiling configurations for each possible input matrix shape, builds a hash table storing these input-optimal tiling pairs, and compiles their code implementations for standby.
At runtime, ATMM adaptively selects the optimal tiling configuration in the hash table, according to the input shapes of both concurrent requests and invoked LoRA adapters, then executes the corresponding code implementation in an extreme efficiency.

\textit{Flexible LoRA adapters orchestration.}
For diverse requirements of vision applications, we propose an orchestrator (\S\ref{sec:LoRAOrchestration}) that efficiently and flexibly orchestrates LoRA adapters at runtime.
Two tools are developed to facilitate high efficiency. 
A switcher leverages ATMM and unified memory management to enable swift inference mode switch and LoRA adapters swap, and a mixture inference mode, deLoRA,  
mitigates the starvation. 
Using the above tools, we design an algorithm to dynamically switch between three inference modes, schedule requests, and manage LoRA adapters to satisfy the performance requirement of each application.


We summarize our key contributions as follows:

\scalebox{0.8}{$\bullet$} To the best of our knowledge, we are the first to identify and solve the key problems in empowering vision applications with LoRA LMM.

\scalebox{0.8}{$\bullet$} We prototype VaLoRA\footnote{Our code is available at \url{https://github.com/mi150/VaLoRA}} that enables accurate, efficient, and flexible LoRA LMM serving for vision applications, which involves accuracy-aware LoRA adapter generation, adaptive-tiling LoRA adapters batching, and efficient and flexible LoRA adapters orchestration.

\scalebox{0.8}{$\bullet$} We implement \name and conduct evaluations for \tr{five} popular analytical tasks with real-world trace on three LMMs. 
Experimental results show that \name \tr{achieves 24-62\% accuracy improvement compared to the original LMMs, and 20-89\% latency} 
reduction compared to the state-of-the-art methods. 


\textbf{This work does not raise any ethical issues.}

\section{Background}\label{sec:backgound}

\begin{figure}[t]
\centering
    \includegraphics[width=0.45\textwidth]{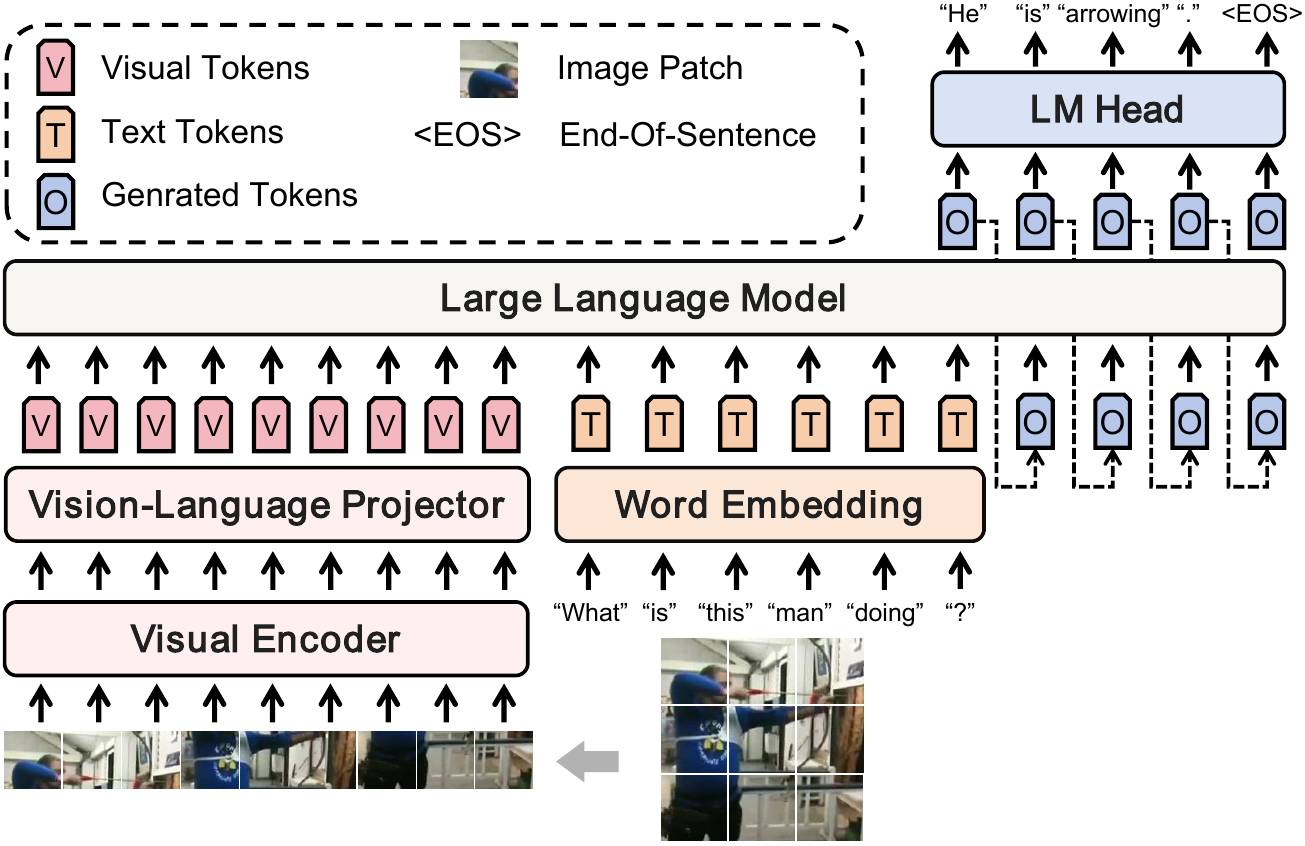}
    \caption{Illustration of LMM inference. Qwen-VL-7B~\cite{QwenVL} generates the right action recognition answer to a piece of data from UCF-101 dataset~\cite{soomro2012ucf101} and the corresponding prompt.}
    \label{fig:LMM}
    \vspace{-1em}
\end{figure}

\noindent
\textbf{Vision applications} in today exploit AI technology to process images or videos in RGB spaces~\cite{zhang2018awstream, khani2023recl, Romil2022Ekya}.
In video analytics, for instance, multiple DNNs that are well-trained on domain-specific datasets separately take care of one target task and together serve the application well~\cite{kang2022viva, kang2017noscope, yu2024vqpy}.
However, the limited capabilities of small models hinder the development of vision applications.
Current applications yet stay on the simple combination of vision tasks such as image classification~\cite{yuan2022PacketGame}, vehicle counting~\cite{li2020reducto}, and target detection~\cite{du2020server}.
With the natural language interface inherited from LLM, LMM can enrich future vision applications.
For example, serving by LMM, the police officer can find the right target when only given a text-described query such as ``A boy wearing a red sweater lost at the corner''.
Therefore, this paper tries to empower vision tasks with LoRA LMM and enrich future vision applications.


\noindent
\textbf{Large multimodal models (LMMs)} aim to achieve stronger general intelligence via extending LLMs with multimodal inputs. 
Since more than 80\% of human beings’ perception and activities are mediated through vision~\cite{thomas2024vision}, it is natural to start the exploration by equipping LLMs with ``eyes''.
By introducing a visual receptor, comprised of a \textit{visual encoder} (\eg ViT~\cite{radford2021learning}) and a \textit{vision-language projector} (\eg Q-former~\cite{li2023blip}), LMMs power LLMs with visual capacity. 
Fig. \ref{fig:LMM} illustrates the inference procedure of LMMs.
Given an image input and its prompt, the visual encoder splits the image into small patches (\eg 14$\times$14 pixels block) and extracts the visual features of each. 
The vision-language projector then converts patches' visual features into visual tokens and feeds into LLM~\cite{achiam2023gpt4, chowdhery2023palm, touvron2023llama2, jiang2023mistral}, together with the embedded text tokens from the prompt, to generate the answer. 
LLM maps the input into high-level features and predicts the probability distribution of the next token with the \textit{language modeling (LM) head} 
in an autoregressive pattern \cite{kwon2023efficient, yu2022orca}. 
As the dotted arrows depicted in Fig. \ref{fig:LMM}, it iteratively generates tokens with input tokens and previous output tokens once a time, until an end-of-sentence (\texttt{<EOS>}) token is emitted. 


\begin{figure}[t]
\centering
    \subfigure[Unmerge mode.] {
    \centering     
     \label{fig:Alongside}     
    {\includegraphics[width=0.23\textwidth]{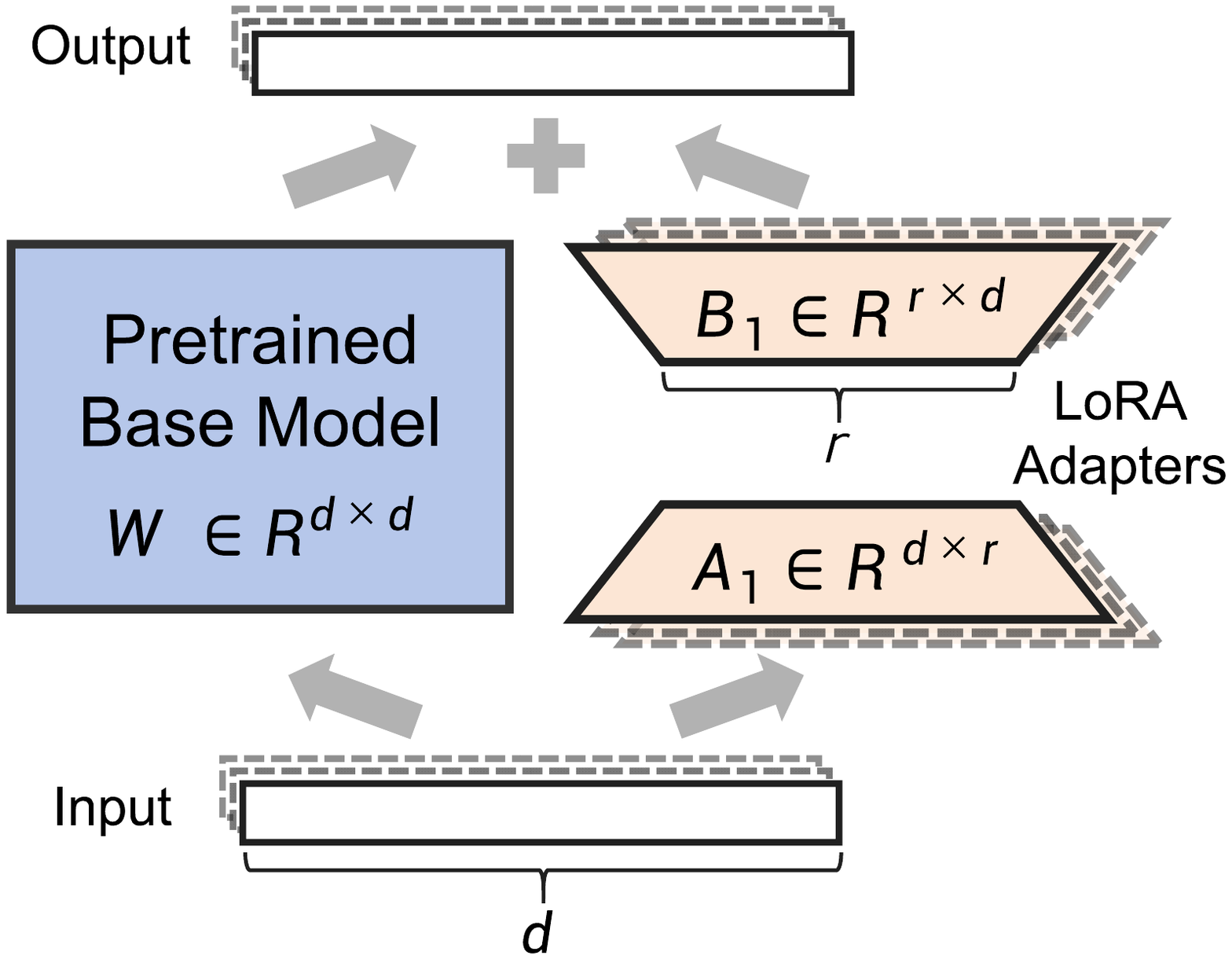}}
    }
    \subfigure[Merge mode.] {
    \label{fig:Merging}
     \centering
    {\includegraphics[width=0.15\textwidth]{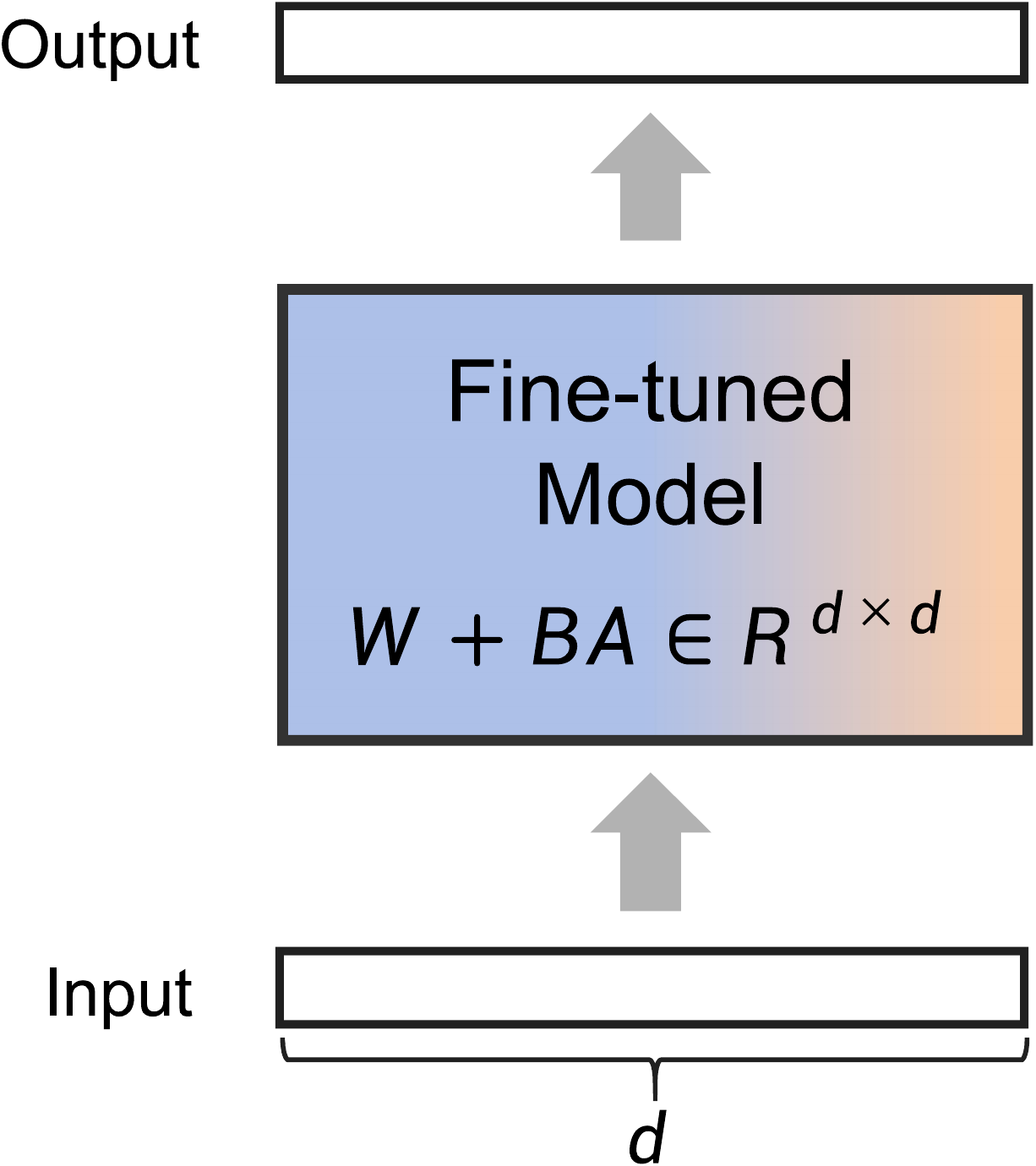}}
    }
    \caption{LoRA model inference. (a) Unmerge mode supports computing multiple different LoRA adapters in a batch. $A_1$ and $B_1$ constitute LoRA adapter \#1. (b) Merge mode supports no-extra-delay inference but only one adapter at once.}
    \label{fig:LoRAInference}
    \vspace{-1em}
\end{figure}


\noindent
\textbf{Low-rank adaptation (LoRA)}~\cite{hu2021lora, dettmers2024qlora, chavan2023one} is a widely used parameter-efficient fine-tuning method~\cite{peft} 
to integrate the external knowledge into large models.
The LoRA adapter is a small number of trainable parameters to learn this knowledge, typically placed in attention layers~\cite{hu2021lora}.
The core of LoRA, as illustrated in Fig. \ref{fig:Alongside}, is to represent each weight update $\Delta W$ as a product of two matrices, $A$ and $B$, with dimensions $r\times d$ and $d\times r$, respectively. 
Their ranks are much smaller than the base model weight $W$, with dimensions $d\times d$.
This makes sense since the low intrinsic rank phenomenon of weight updates in large models~\cite{hu2021lora}.
When fine-tuning, LoRA adapters only update $A$ and $B$ while keeping $W$ frozen. 
At inference, the computation of all LoRA adapters can be placed bypass, as shown in Fig. \ref{fig:Alongside}, and only adds the results onto the output.  
Or merge one multiplied matrix $B\times A$ (\ie $\Delta W$) into base model $W$, which keeps computation overhead the same as the base model in Fig. \ref{fig:Merging}.


\noindent
\textbf{LoRA models serving system}~\cite{chen2024punica, sheng2023slora, wu2024dlora} targets to improve LoRA model inference efficiency. 
In unmerged inference, the core characteristic is that computing two small matrices of the adapter underutilizes GPU computing units, leading to unnecessary delay and resource waste.
To tackle this, Punica~\cite{chen2024punica} and S-LoRA~\cite{sheng2023slora} batch multiple heterogeneous adapters, as the cascade LoRA adapters illustrated in Fig. \ref{fig:Alongside}, and compute them within one single custom CUDA kernel.
This method boosts the system throughput indeed but causes significant additional overhead (more in \S \ref{sec:challenges}).
dLoRA~\cite{wu2024dlora} merges the most accessed LoRA adapter into the base model and switches to unmerge if necessary.  
However, its mode switch causes an unacceptable time cost (more in \S \ref{sec:challenges}).
Besides, like Punica and S-LoRA, dLoRA's unmerged inference also fails to address the large number of extra computation overhead.

\section{Motivation and Challenges}
This section explores two questions: 
(1) What benefits can LoRA LMM bring to vision applications (\S \ref{sec:opportunity})?
(2) What challenges must be tackled when empowering vision applications with LoRA LMM (\S \ref{sec:challenges})?

\begin{figure}[t]
\centering
    \subfigure[Zero-shot Grounding results on data \#38 in Aircraft dataset~\cite{Aircraft}.] {
    \centering     
     \label{fig:ZeroshotGrounding}     
    {\includegraphics[width=0.195\textwidth]{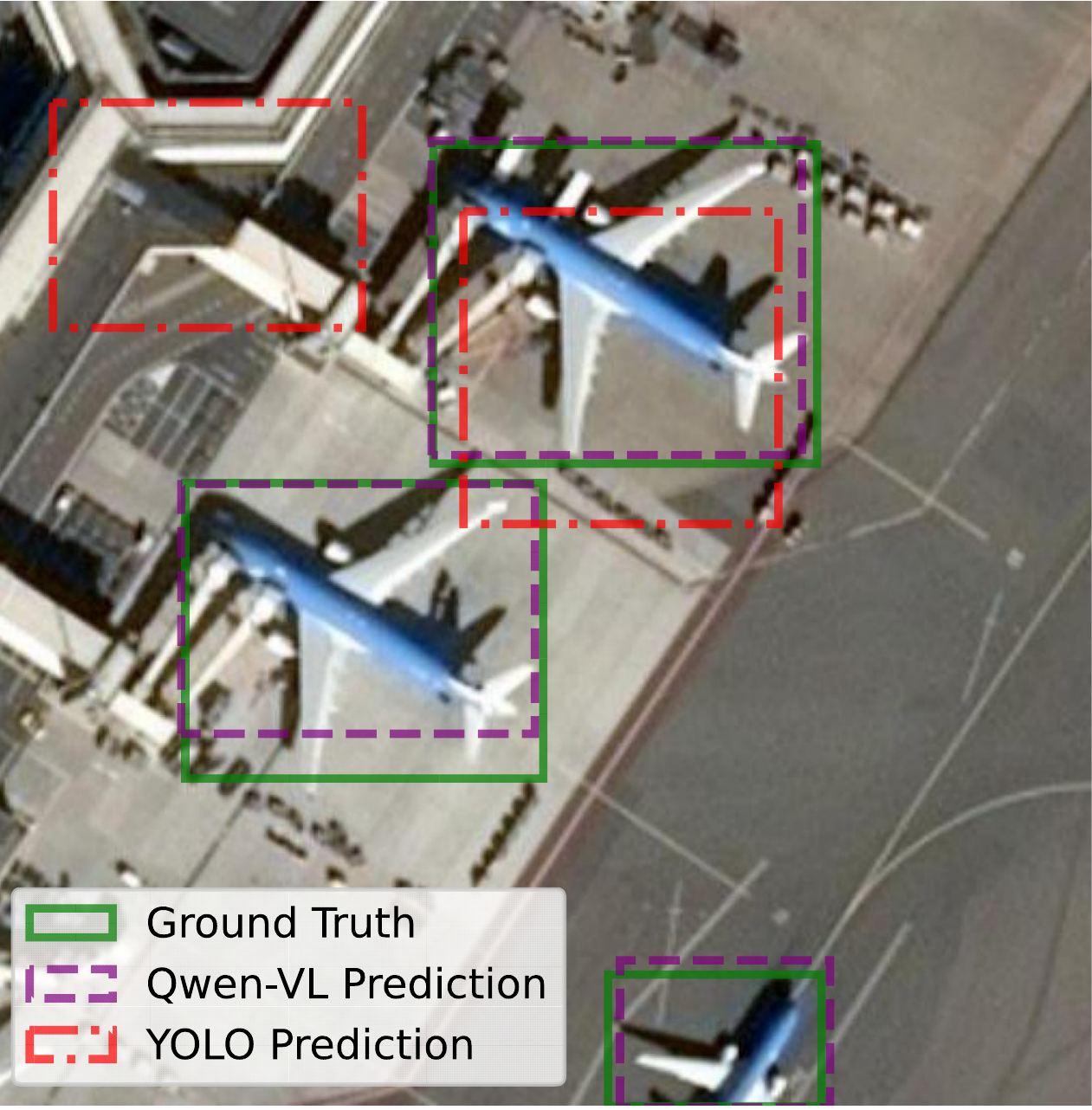}}
    }
    \hspace{0.5em}
    \subfigure[Visualization of data \#133279 in Visual Question Answering~\cite{vqa2}.] {
    \label{fig:GeneralVQA}
     \centering
     \raisebox{0.05cm}
    {\includegraphics[width=0.19\textwidth]{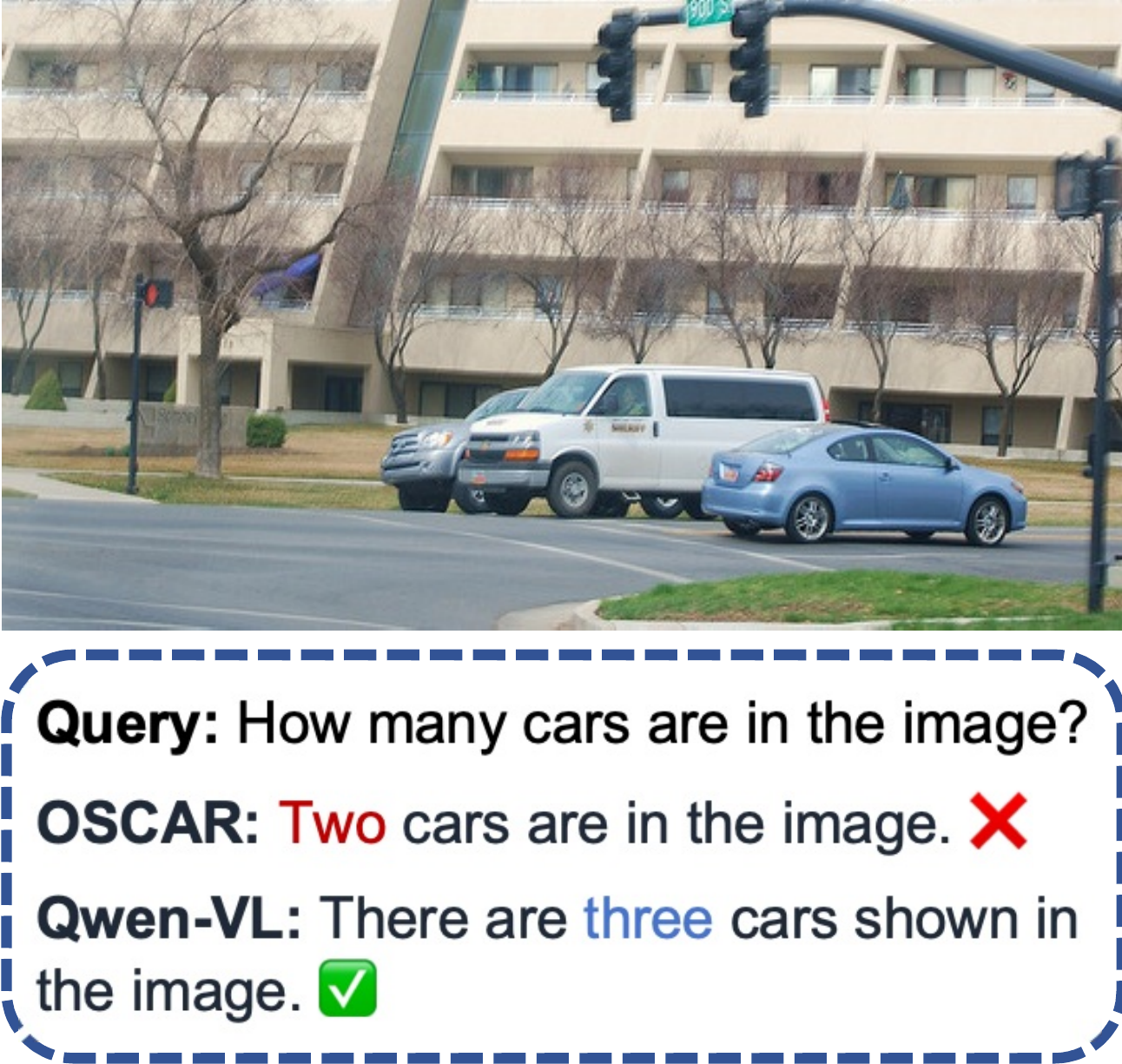}}
    }
    \vspace{-1.5em}
    \caption{The potential of LMM. (a) To ground the airplanes in remote sensing view in zero-shot, LMM Qwen-VL, in general, delivers 67.2\% accuracy \vs the 18.3\% of YOLO~\cite{glenn2021YOLOV5}. (b) In VQA, Qwen-VL yields 78.8\% accuracy \vs the 73.3\% of OSCAR~\cite{li2020oscar}.
    }
    \label{fig:Opportunity}
    \vspace{-1.em}
\end{figure}

\subsection{Potential Benefits from LoRA LMM}\label{sec:opportunity}
\textbf{LMMs offers state-of-the-art performance on many complex vision tasks.}
To demonstrate it, we take zero-shot grounding and visual question answering as examples, and conducted experiments on Aircraft~\cite{Aircraft} and VQAv2~\cite{vqa2} datasets with Qwen-VL-7B~\cite{QwenVL} (as the LMM), YOLO~\cite{glenn2021YOLOV5} and OSCAR~\cite{li2020oscar} (as the baseline small models).
Aircraft contains 103 remote sensing images that are not pre-trained on Qwen-VL and YOLO, being the zero-shot test; VQAv2 is the most popular visual question answering dataset which includes text and visual modalities, to test the multi-modal ability.
Fig. \ref{fig:Opportunity} shows that, with the solid linguistic and reasoning capabilities inherited from LLMs, Qwen-VL greatly outperforms small models.
It delivers 48.9\% higher F1-score in zero-shot grounding than YOLO.
Fig. \ref{fig:ZeroshotGrounding} visualizes the results on data \#38, where Qwen-VL bounds more accurate boxes than YOLO.
For multi-modal tasks, Qwen-VL achieves 78.8\% accuracy, being 7.5\% higher than OSCAR.
Fig. \ref{fig:GeneralVQA} exemplifies a typical vehicle counting task in video analytics applications, only Qwen-VL generates the correct answer.

\begin{figure}[t]
\centering
     \centering
    \includegraphics[width=0.85\linewidth]{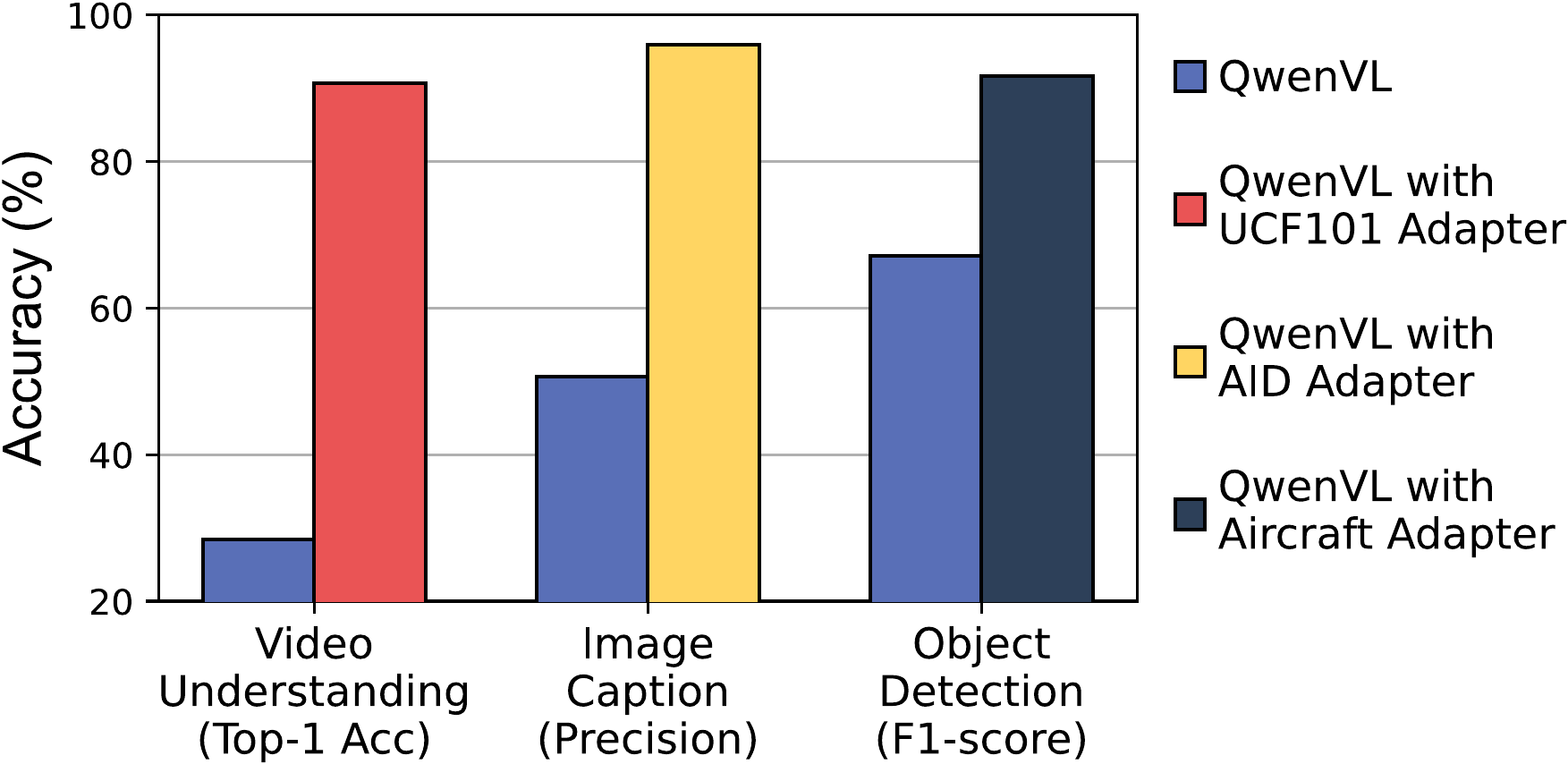}
    \caption{LoRA adapters with domain-specific knowledge improve the Qwen-VL's accuracy on target tasks.}
    \label{fig:FT_dc}
    \vspace{-1em}
\end{figure}

\textbf{With external knowledge from LoRA adapters, LMM attains remarkable accuracy gain on domain-specific tasks.}
To investigate the accuracy improvement from external knowledge, 
we fine-tune three LoRA adapters for image classification, object detection, and video classification, respectively, on external datasets, AID~\cite{AID}, Aircraft~\cite{Aircraft}, and UCF101~\cite{UCF101}.
Fig. \ref{fig:FT_dc} shows the results.
With fine-tuned LoRA adapters, Qwen-VL receives 45.2\%, 24.5\%, and 62.2\% accuracy gains on three domain-specific task, respectively.
Note that we only validated the potential gain without fully exploring advanced training techniques like data enhancement~\cite{VideoEnhancement}. 
Nevertheless, based on current results, we believe these techniques could further improve accuracy in future work.

\textbf{LoRA LMM enables more flexible serving.}
Today's vision applications are served by many domain-specific small models~\cite{Romil2022Ekya, khani2023recl}.
When invoked, the specific model is loaded into GPU and swapped out to main memory after execution~\cite{shen2019nexus, bai2020pipeswitch}.
This swapping method allows model execution in a large batch but incurs inevitable transmission costs. 
LoRA adapters
usually have fewer parameters than small models (more in \S\ref{sec:Switch}).
Swapping them while keeping LMM in GPU provides a more flexible serving.
In our test, swapping an adapter saves 97\% delay than OSCAR, 15ms \vs 520ms, and 86\% of YOLO's 110ms.

\subsection{Challenges of Empowering Vision Applications with LoRA LMM
}\label{sec:challenges}
To empower diverse vision applications, the LoRA LMM system must offer accurate and efficient responses to the vision tasks involved and meet the distinct application-specified performance requirements.
To this, we face three challenges.

\begin{figure}[t]
\centering
     \centering
    \includegraphics[width=0.75\linewidth]{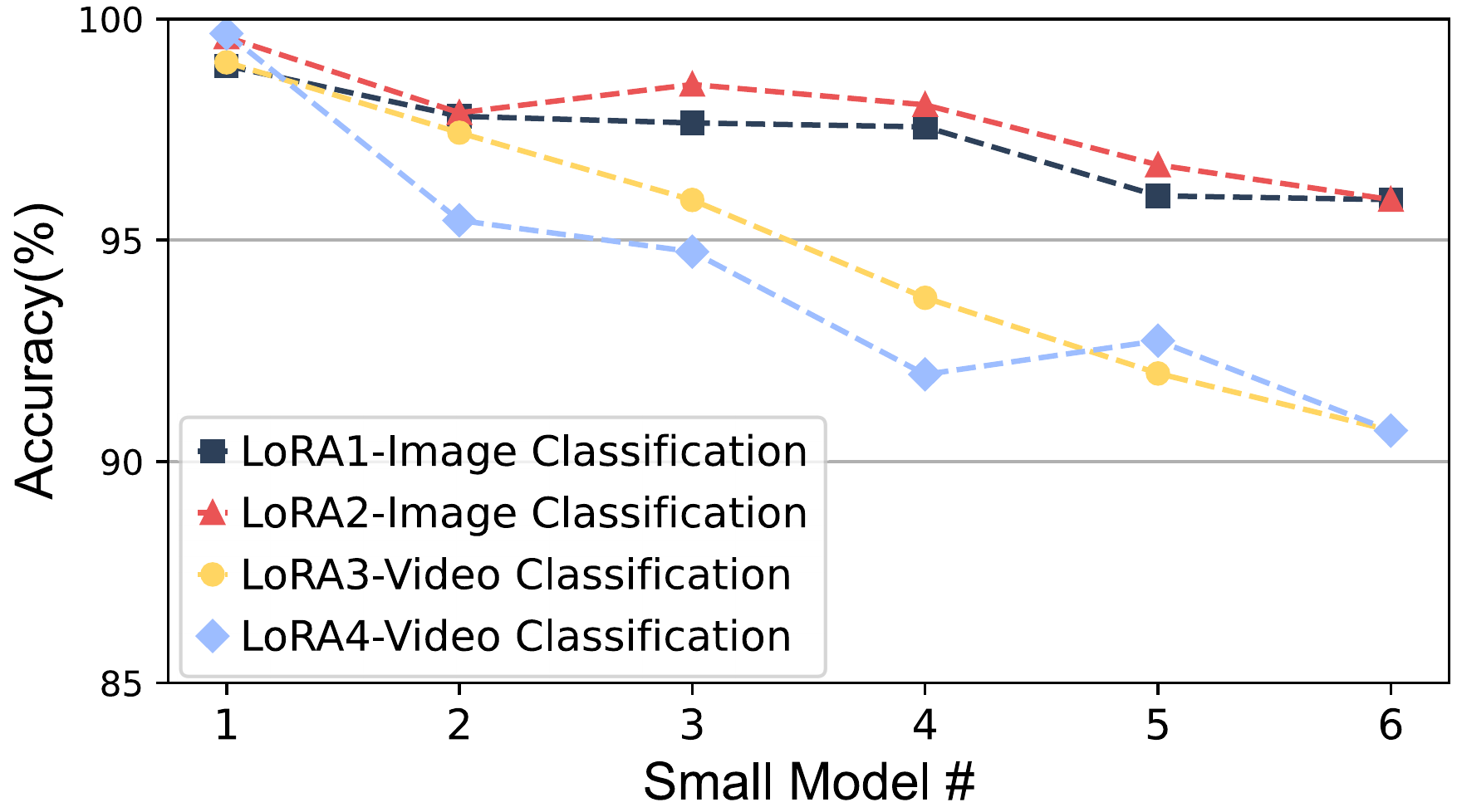}
    \caption{Accuracy decreases when fusing knowledge from multiple domain-specific small models into one single LoRA. 
    The trend varies regarding vision tasks.  
    }
    \label{fig:task_num}
    \vspace{-1em}
\end{figure}

\textbf{C1: Limited capacity of LoRA adapter.}
Integrating external knowledge from existing small models or specific datasets into LMM is essential to generate accurate results.
Parameter-efficient fine-tuning LoRA adapters show promise. 
However, it is challenging because the LoRA adapter has only limited capacity and varies on the vision tasks.
Fig. \ref{fig:task_num} demonstrates this by fusing external knowledge from different numbers of small models on diverse tasks into a single LoRA adapter (experiment setup details in \S \ref{sec:ExperimentalSetup}).
Training a separate adapter for each small model consistently achieves high accuracy but results in significant adapter capacity waste, while fusing too many small models into one adapter incurs significant accuracy degradation.
For example, the LoRA adapter that fuses six image classification models in Fig. \ref{fig:task_num} retains over 95\% accuracy, while fusing six video classification models decreases remarkable accuracy.

\textbf{C2: Inefficient concurrent requests batching.}
One vision application often involves multiple small models~\cite{jiang2018chameleon, du2020server}. 
Hence, serving multiple vision applications with LMM very likely leads to the simultaneous invocation of multiple heterogeneous LoRA adapters. 
However, current LoRA model inference systems struggle to process them efficiently, particularly under high concurrency. 
To demonstrate this, we measure three state-of-the-art systems.
\emph{Punica}~\cite{chen2024punica} and \emph{S-LoRA}~\cite{sheng2023slora} customize CUDA operator, 
respectively, to 
batch heterogeneous LoRA adapters computation in unmerge mode (more details in \S \ref{sec:ATMM}), while \emph{dLoRA}~\cite{wu2024dlora} calls PyTorch operator \texttt{Einsum}\footnote{
Einsum uses Einstein summation convention~\cite{Einstein1922}, describing tensor index operations in a concise string form, for efficient LoRA adapter batching.}~\cite{torcheinsum}.
The experimental workload randomly generates 2-4 requests ranging from 128 to 1024 length of input tokens per second, and we repeat 1,000 times to measure the latency.
For fairness, all experiments run on a server device equipped with NVIDIA A100 GPU~\cite{A100} via PEFT~\cite{peft} framework, and use Qwen-VL-7B~\cite{QwenVL} as the base model.

\begin{figure}[t]
\centering
\setlength{\abovecaptionskip}{5pt}
  \begin{minipage}[b]{0.49\linewidth}
  \centering
  {\includegraphics[width=0.9\linewidth]{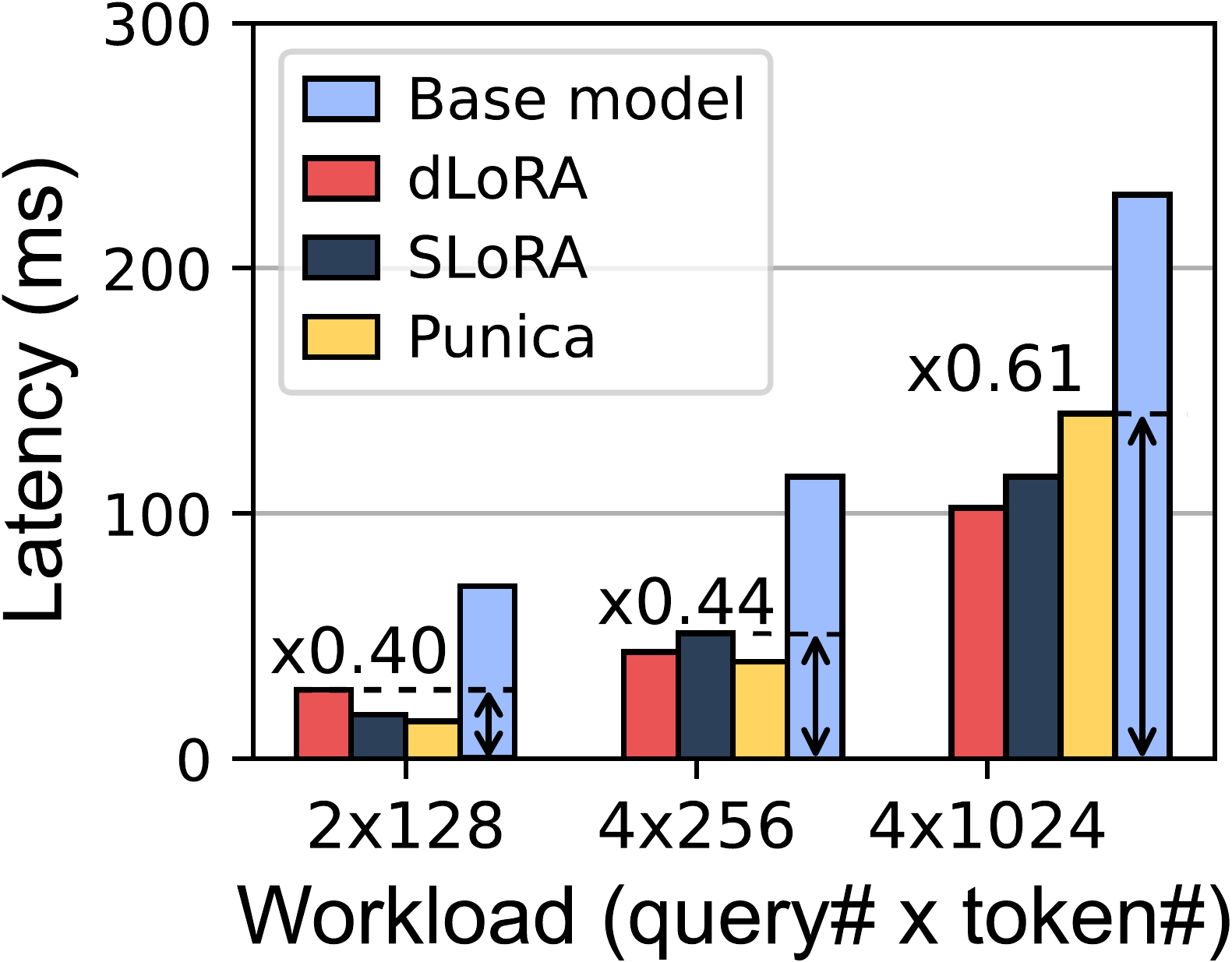}}
    \caption{Unmerged inference causes 27-140ms extra latency equivalent to 40-61\% of base model inference time.}
    \label{fig:OPTile}
    \end{minipage}
\hfill
  \begin{minipage}[b]{0.48\linewidth}
  \centering
     \raisebox{0.05cm}
    {\includegraphics[width=0.9\linewidth]{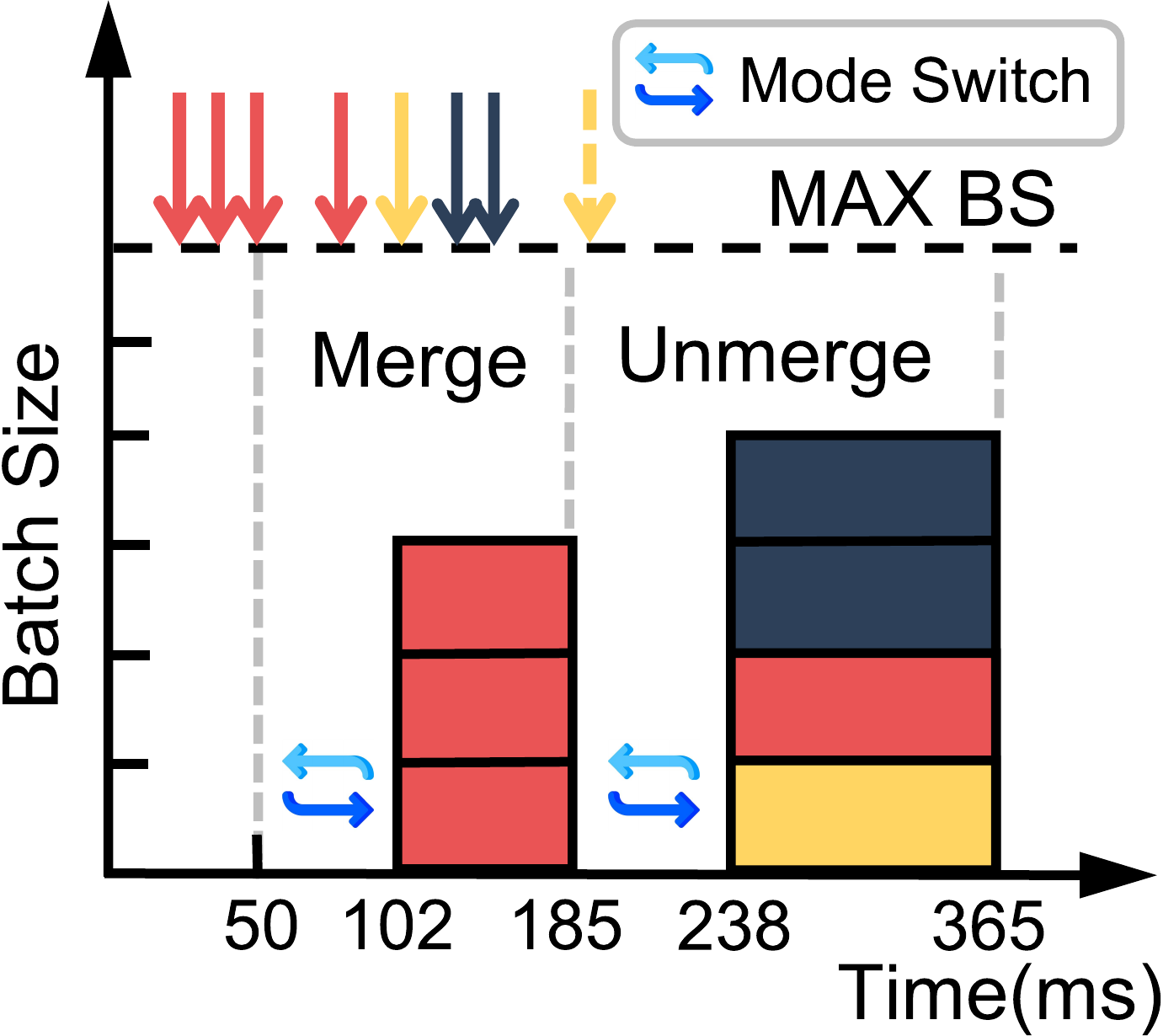}}
    \caption{Mode switch alone costs 53ms, occupying 64\% of merged inference time of three 256-tokens requests.}
      \label{fig:SwitchCost}
  \end{minipage}
  \vspace{-1em}
\end{figure}

Fig. \ref{fig:OPTile} plots the results.
The bars of dLoRA, S-LoRA, and Punica denote their extra latency than the merged inference, \eg $latency_{dLoRA}-latency_{merge}$.
The bar of the base model denotes its time cost under the same workload. 
Unmerged inference yields up to 140ms additional latency when serving four 1024-token requests. 
This unnecessary waste is sufficient for the base model to perform 4x256 inference, with resources to spare!
The reason for such a high cost is two-fold. 
1) Inherently, additional overhead stems from two additional matrix multiplications and one additional matrix addition per layer as depicted in Fig. \ref{fig:Alongside}; 
meanwhile, these computations run in parallel with the base model computations, each layer requires additional CUDA kernel context operations at each layer.
2) Upon concrete implementations, all three LoRA adapters batching operators fail to fully utilize computing units in GPU, due to the significant padding or ill-considered matrix tiling (more in \S \ref{sec:ATMM}).




\textbf{C3: Inflexible LoRA adapters orchestration.}
To cope with the distinct performance requirements specified by vision applications, 
an orchestration that can carefully manage LoRA adapters and flexibly schedule application requests is necessary.
We believe the inference mode switch like dLoRA is promising, yet it falls significantly short of efficiency.
Its mode switch yields unacceptable overhead.
Fig. \ref{fig:SwitchCost} illustrates a real scheduling state and mode switch latency for two consecutive inference slots of dLoRA.
In this case, dLoRA serves 8 requests, each with an input length of 256, in a first-come-first-service manner.
In the first slot, dLoRA serves requests 1-3 in merge mode using the same LoRA adapter, and the heterogeneous requests 4-7 are processed in an unmerged mode in the following slot. 
A mode switch delay of over 53 ms makes the last request (the dotted arrow in Fig.\ref{fig:SwitchCost}) have to wait 165ms until the next inference slot begins.
This significant cost stems from 1) unnecessary memory copy of LoRA matrices due to dLoRA's inefficient memory management, and 
\tb{2) the substantial overhead of LoRA matrices $\Delta W$ computation, by matrix multiplication $A\times B$, then added (merge) or subtracted (unmerge) $\Delta W$ onto or from the base model, by invoking \texttt{torch.addmm}, per layer.}
Conceivably, if the mode switch can be reduced to <10ms (as this paper achieved in \S \ref{sec:Switch}), the average response time of Fig.\ref{fig:SwitchCost} case can save 45ms, with the last request only need to wait <80ms.



\section{\name Design}
\name is an end-to-end system that empowers diverse vision tasks and enriches vision applications with LoRA LMM by addressing the above challenges. 
We first provide an overview of \name's operation, then describe three core techniques it leverages. 
\subsection{System Overview}
\begin{figure}[t]
\setlength{\abovecaptionskip}{3pt}
\setlength{\belowcaptionskip}{3pt}
\centering
\includegraphics[width=0.95\linewidth]{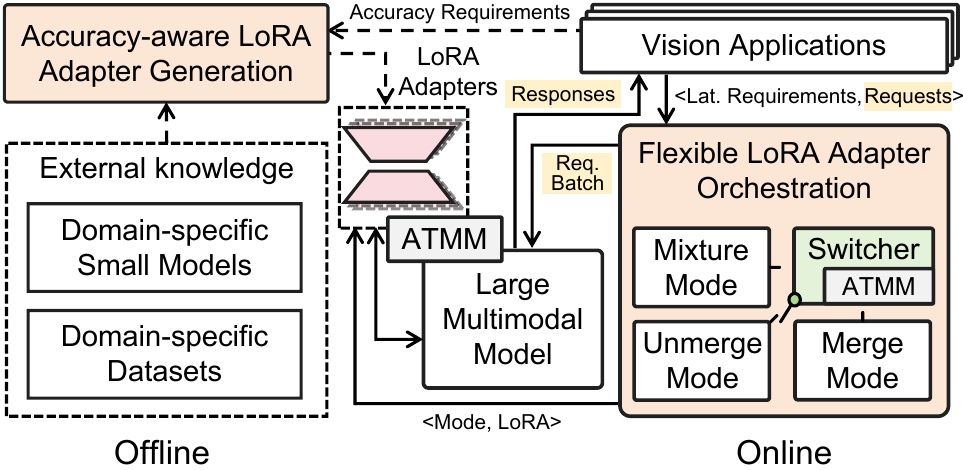}
     \caption{VaLoRA overview.}
    \label{fig:VaLoRA}
    \vspace{-1em}
\end{figure}

\name includes two phases.
During the offline phase, the accuracy-aware LoRA adapter generation approach takes the external knowledge, from the existing domain-specific small models or datasets, as well as the accuracy requirements specified by vision applications (as the dotted arrows plotted in Fig.\ref{fig:VaLoRA}), to generate the minimum number of LoRA adapters (\S\ref{sec:LoRAGeneration}). 
The generated LoRA adapters are rich in domain-specific knowledge that can output accurate responses to tasks involved in vision applications.

During the online phase, the flexible LoRA adapter orchestration ingests the requests from vision applications,
organizes them into batches, chooses the inference mode, and orchestrates their corresponding LoRA adapters, to minimize the average response latency while guaranteeing each vision application's latency constraint (\S\ref{sec:LoRAOrchestration}).
Each request batch is delivered to the corresponding adapters and LMM and inferred in the chosen mode.
The LoRA adapter batching and inference mode switcher are implemented with ATMM, the adaptive-tilling matrix multiplication operator (\S\ref{sec:LoRABatching}), achieving high efficiency.




\subsection{Accuracy-aware LoRA Adapter Generation}\label{sec:LoRAGeneration}
To offer accurate results on domain-specific tasks, we propose the accuracy-aware LoRA adapter generation consisting of an accuracy-aware knowledge-fusion algorithm and 
the vision task head. 

\subsubsection{Accuracy-aware knowledge-fusion algorithm.}
To make it easy to manage at runtime, we aim to integrate external knowledge into the fewest LoRA adapters without violating the accuracy requirements of any vision tasks.  
To this end, the training method must account for the limited capacity of the LoRA adapter and the complex accuracy variations arising from the knowledge fusion. 

This problem is highly challenging. 
Suppose we have an oracle who knows the accuracy of a LoRA adapter that fused arbitrary knowledge combinations in advance.
Then, the problem can be formulated as a \textit{constrained bin-packing problem}~\cite{garey1981approximation}, where the objective is to pack the knowledge into the minimum number of LoRA adapter bins, ensuring each adapter maintains each vision task's accuracy beyond the requirement. 
However, this relaxed variant is NP-hard, and unfortunately, such an oracle does not exist.

\begin{figure}[t]
\centering
\includegraphics[width=1\linewidth]{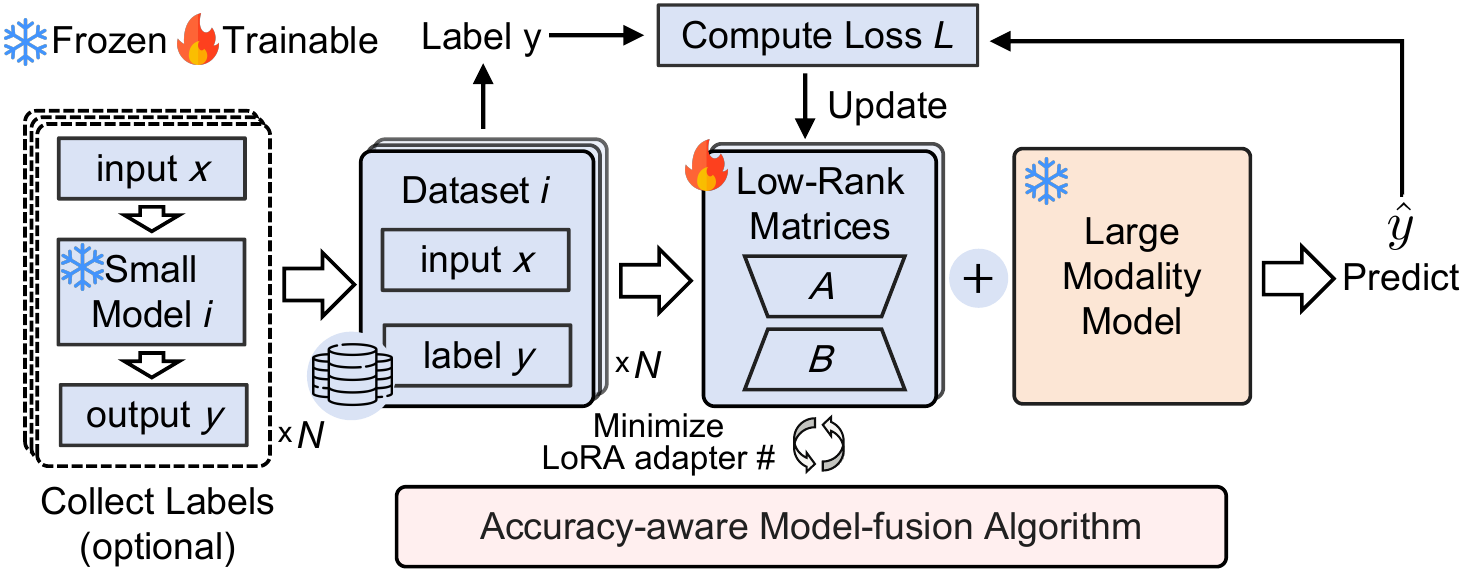}
     \caption{Accuracy-aware LoRA generation integrates the domain-specific knowledge into LoRA adapters with the proposed accuracy-aware knowledge-fusion algorithm.}
    \label{fig:LoRAGeneration}
    \vspace{-0.5em}
\end{figure}

\tb{To solve the original problem, we propose a simple and easy-to-implement heuristic algorithm, the \textit{accuracy-aware knowledge-fusion algorithm}, to determine which knowledge fuses into one LoRA adapter.}
It first collects the dataset, as illustrated in Fig. \ref{fig:LoRAGeneration}, by executing representative data on every existing domain-specific small model; if applications provide the datasets, we directly use them. 
After that, the training process is a standard supervised learning pipeline that computes the cross-entropy loss, $L=CE(y,\hat{y})$, for parameter update.
The knowledge fusion employs greedy and accuracy-aware heuristics.
It begins with a random dataset and sequentially uses each dataset to train the LoRA adapter until its accuracy on a specific task falls below the required threshold.
If this occurs, the adapter's weights are rolled back, and a new adapter is initialized to learn from the most recent dataset.
The worst case of such a method may generate one LoRA adapter for each dataset, but in our practical experiments, every LoRA adapter fuses 4 domains of knowledge (datasets) on average.

\begin{figure}[t]
\centering
     \centering
    \includegraphics[width=1\linewidth]{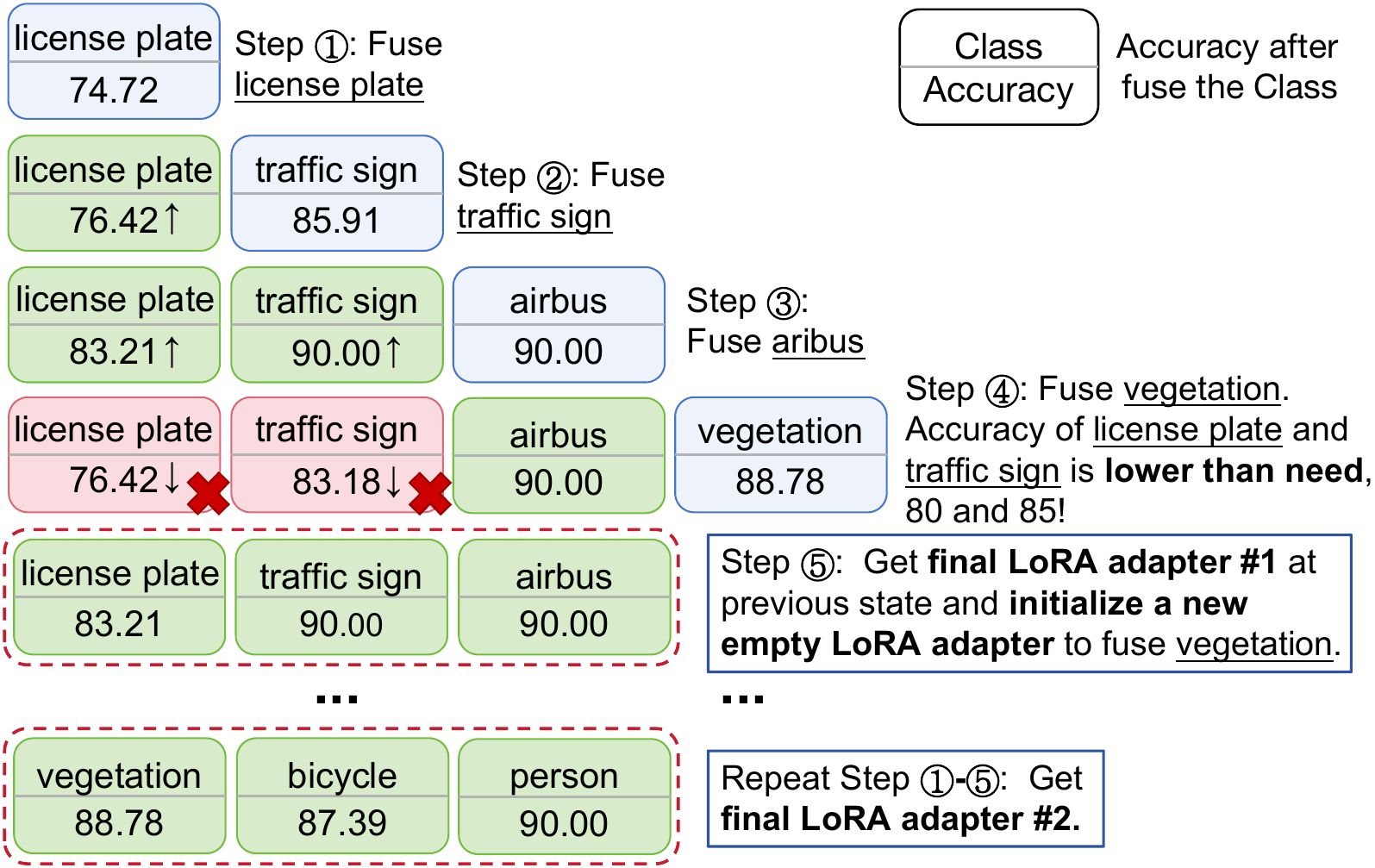}
    \caption{\tb{An example of fusing the knowledge from six object detection models, each detects one class of object, with the accuracy-aware knowledge-fusion algorithm.
    }}
    \label{fig:ModelFusion}
    \vspace{-1em}
\end{figure}

Fig. \ref{fig:ModelFusion} illustrates an example of integrating the knowledge of six object detection models, each detecting one class of target, into LoRA adapters.
When fusing the vegetation-detect model (Step \ding{175}), the accuracy of LoRA adapter 1 fails on the license plate, needs above 80\%, and traffic sign detection, 85\%.
Hence, the algorithm returns the LoRA adapter 1 of the previous state and initializes LoRA adapter 2 (Step \ding{176}).
As the vegetation, bicycle, and person datasets are fused into LoRA adapter 2 without accuracy violation, we get the second LoRA adapter (Step \ding{177}).
As LMM has powerful learning abilities, the training procedure costs only 25 minutes in this example. 
We leave the exploration of this to the future.






\subsubsection{Vision task head.}\label{sec:VAHead}
\begin{figure}[t]
\centering
{\includegraphics[width=0.95\linewidth]{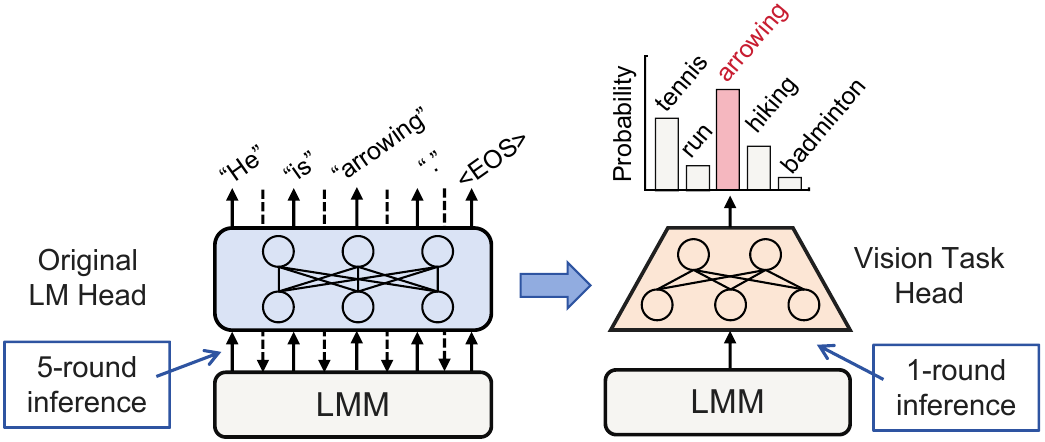}}
    \caption{\tb{Vision task head. Replacing the original language modeling head in LMM with vision task head reduces 4 rounds of LMM inference on the action recognition task of Fig. \ref{fig:LMM}.}}
    \label{fig:VsionTaskHead}
    \vspace{-1em}
\end{figure}

To reduce the inference latency, we design the vision task head. 
It is designed as a trainable linear layer as a part of the LoRA adapter, as shown in Fig. \ref{fig:VsionTaskHead}, to predict task-specific results based on the output features of LMM. 
\tb{Vision task heads can be flexibly customized to various vision tasks during the LoRA adapter training\footnote{As the vision task head is served as a part of the LoRA adapter, it is included in the cost of training a LoRA.}, \eg action recognition head in Fig. \ref{fig:VsionTaskHead}, provided the fusing knowledge is from the same task type. 
Fig. \ref{fig:VsionTaskHead} compares doing action recognition with the original language modeling (LM) head and the vision task head. 
By replacing LM head with the vision task head, LMM saves 4 inference rounds, around 180ms time cost.}
The reason to do so is that the outputs of a large portion of vision tasks are a limited discrete set of candidate options, such as the number of vehicle counts~\cite{li2020reducto}, classes of action recognition~\cite{yuan2022PacketGame}, and binary query for a specific target on image or video~\cite{yi2020eagleeye}. 

\tb{We retain the LM head for vision applications that need the natural language interface. 
For example, when video query applications ask for ``A boy wearing a red sweater lost at the corner'' and specify the person detection, \name will invoke the corresponding LoRA adapter containing a detection head for efficient response\footnote{Some work, \eg task automation \cite{wen2024autodroid, li2024personal} and dynamic LoRA \cite{feng2024mixture}, automatically identify adapters from queries. They are orthogonal to \name.}.
By using the LoRA adapters batching operator, ATMM, in the next section, these different vision task heads can be executed concurrently, as well as in parallel with the LM head.}

\begin{table}[t] 
\footnotesize
\centering 
\hspace{-0.5em}
\begin{tabularx}{0.485\textwidth}{c|c c}
\toprule 
\makecell{\textbf{Tiling} \\ \textbf{Configuration}} & \makecell{\textbf{Input~\ding{172}} \\ (256$\times$4096, 4096$\times$32)} & \makecell{\textbf{Input~\ding{173}} \\ (8192$\times$4096, 4096$\times$128)} \\
\midrule 
\makecell{Punica~\cite{chen2024punica}\\ {\footnotesize(16, 64, 64, 16, 16, 64)}}  & \makecell{0.087ms\\Low SM Utilization} & \hspace{-1em} \makecell{0.19ms\\Freq. Global Mem. Access}\\
\midrule 
\makecell{Config~\ding{172} \\ {\footnotesize(64, 32, 32, 32, 32, 32)}}  & \makecell{\textbf{0.07ms}} & \hspace{-1em} \makecell{0.12ms\\Freq. Global Mem. Access}\\
\midrule 
\makecell{Config~\ding{173} \\ {\footnotesize(64, 64, 64, 32, 64, 64)}}  & \makecell{0.13ms\\Low SM Utilization} & \hspace{-1em}\makecell{\textbf{0.10ms}}\\
\bottomrule 
\end{tabularx}
\caption{
Adaptive tiling saves remarkable computing time compared to static-tiling Punica.
Configuration (a,b,c,d,e,f) indicates the thread block tile a$\times$b, b$\times$c, and warp tile d$\times$e, e$\times$f.
We omit thread tile to save space.
Input (u$\times$v, s$\times$t) indicates input matrix shapes.
The text under latency explains why this configuration performs poorly under corresponding input.
} 
\label{tab:AdaptiveTiling}
\vspace{-3em}
\end{table}

\subsection{Adaptive-tiling LoRA Adapters Batching}\label{sec:LoRABatching}
Serving vision applications with LMM very likely invokes heterogeneous LoRA adapters concurrently.
To compute them in high performance, we propose an Adaptive-Tiling Matrix Multiplication operator, ATMM, enabling efficient unmerged inference, inference mode switching (\S\ref{sec:Switch}), and mixture inference mode (\S\ref{sec:deLoRA}).




\subsubsection{ATMM: Adaptive-tiling matrix multiplication operator.}\label{sec:ATMM}
Directly batching heterogeneous adapters computation upon standard kernels, as batched GEMM (General Matrix Multiplication) in dLoRA~\cite{wu2024dlora}, is feasible, but yields excessive latency and hardware underutilization. 
This stems from the significant padding arising from the heterogeneity of application request lengths and LoRA adapter ranks.
Hence, a customized kernel 
is necessary.


To motivate our design, we first analyze two existing customized kernels from S-LoRA~\cite{sheng2023slora} and Punica~\cite{chen2024punica}, respectively. 
S-LoRA's kernel utilizes \emph{tiling} technique to avoid the significant padding. 
Unlike batched GEMM, which pads heterogeneous input matrices to a uniform shape, it splits them into fine-grained blocks and computes the output for each block in parallel on CUDA cores.
Punica's kernel also employs the tiling technique and further enhances efficiency by leveraging the well-developed CUTLASS~\cite{cutlass} library and higher-performance Tensor cores~\cite{tensorcore}.
However, as the motivational study in \S \ref{sec:challenges}, both kernels fail to achieve satisfactory efficiency.
The root cause is their static tiling configurations, which are inadequate to handle diverse input shapes, resulting in underutilized computational resources.

\begin{figure}[t]
\centering
\setlength{\abovecaptionskip}{0pt}
\setlength{\belowcaptionskip}{-5pt}
\vspace{-0.5em}
    \subfigure[Under heavier Input~\ding{173}, Punica's tiling (upper half) yields more tiles, so more costly memory read/write compared to Config \ding{173} (lower half).] {
    \centering     
     \label{fig:CaseStudy1}     
    {\includegraphics[width=1\linewidth]{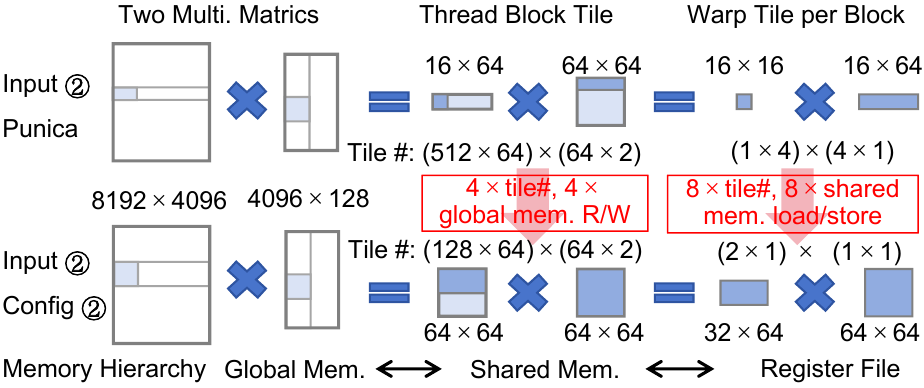}}
    }\vspace{-1em}
    \subfigure[But under lighter Input~\ding{172}, Config~\ding{173} (lower half) cannot deliver good performance because it underutilizes SM since fewer block number.] {
    \label{fig:CaseStudy2}
     \centering
     \vspace{-0.5em}
    {\includegraphics[width=1\linewidth]{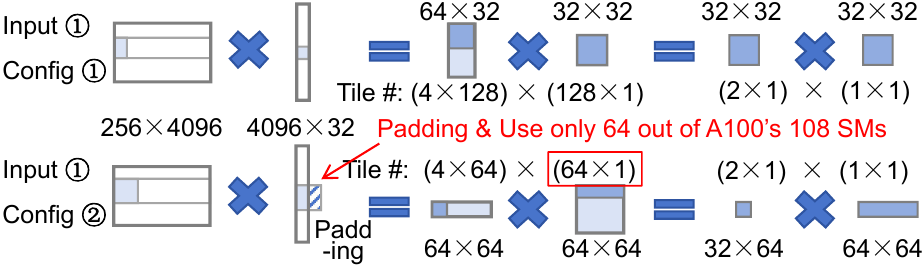}}
    }
    \caption{\tb{Paired comparison of tilling configurations in Table~\ref{tab:AdaptiveTiling}. 
    Following tiling configurations, two multiplied matrices are divided into thread block tiles and further warp tiles. 
    The data in each tile is transferred to their corresponding memory. 
    }}
    \label{fig:ATMM}
    \vspace{-1.5em}
\end{figure}

Our key observation is that the computational efficiency varies significantly with different tiling configurations.
We conduct an experiment with two input shapes and three tiling configurations and present the results in Table \ref{tab:AdaptiveTiling}. 
Under Input~\ding{172} and Input~\ding{173},
three tiling configurations yield the most 1.9$\times$ latency difference. 
This stems from the inappropriate tiling against input shape and hardware architecture, as well as between tiling levels.
\tb{
Fig. \ref{fig:ATMM} compares two configuration pairs in Table \ref{tab:AdaptiveTiling}.
In Fig.\ref{fig:CaseStudy1}, Punica's smaller-tile configuration produces more thread block tiles and warp tiles.
Compared to Config~\ding{173}, it results in more frequent accesses to global and shared memory, so more launching data transfer times incur Punica's 1.9$\times$ latency.
However, larger tile does not always benefit.
In Fig.\ref{fig:CaseStudy2},  Config~\ding{173}'s large tiling incurs Streaming Multiprocessor (SM) under-utilization. 
It can only utilize 64 of 108 SMs in A100, as each thread block tile can only fetch one SM, far less than Config~\ding{172}.
In sum, we need dynamic tiling to achieve efficiency.
}

We propose ATMM, an adaptive-tiling matrix multiplication CUDA kernel that fully utilizes computational resources, achieving efficient heterogeneous LoRA adapters batching.
It has two key design choices. 
(1) Adaptive tiling against input shapes.
Given two input matrices, ATMM retrieves the optimal tiling configuration from the \textit{hash table} (established in \S \ref{sec:SearchAlg}), and following this configuration, divides the matrix into \textit{thread block tiles}, then further into \textit{warp tiles}\footnote{Each warp can be divided into thread tiles in <16,8,16> or <16,8,8> shape.}.
(2) Pipeline data loading and computing. 
After tiling, ATMM transfers the data of each tile to their corresponding memories, the correspondence as shown in Fig.~\ref{fig:CaseStudy1}, and executes computation on CUDA/Tensor cores with the corresponding executable kernel (pre-compiled in \S \ref{sec:SearchAlg}).
To hide data loading latency, ATMM allocates double the space in shared memory and register file for each tile: one for the current tile’s computation and one for prefetching the next tile's data. 
Note that such double buffering refers to the use of dual buffers in shared memory, an on-chip cache (\eg 20MB in A100) independent from GPU memory (\eg 80G in A100). 
It is essentially a cache usage strategy without incurring global memory overhead.
This double buffering is feasible due to each level's uniform tile shape, enabling efficient location of the next tile.

\subsubsection{Profile-based optimal tiling search.}\label{sec:SearchAlg}
Keeping the benefit of adaptive tiling in mind, we aim to search out the optimal tiling configuration for every different input and switch among them at runtime.
This can be done offline, as it is only affected by input shapes rather than their numerical values.
Some prior work~\cite{Roller, Bolt, shi2023welder} though studied the optimal tiling for matrix multiplication, it is still challenging as the complicated GPU thread parallelism mechanism and memory access patterns, and the large search space of input shapes.

To tackle this challenge, we approach the search as a black-box problem and propose a profile-based searching algorithm.
It profiles the execution time for all possible ATMM input matrix shapes with the help of CUTLASS Profiler~\cite{cutlass}, records the optimal tiling configurations in a hash table, and compiles corresponding executable kernels. 
We use the following expert knowledge to reduce the search space. 
(1) From the hardware perspective, architectural characteristics restrict the feasible input shapes, \eg limited memory of each hierarchy, and tiling configurations, \eg every dimension of tile at least 16 and must be powers of two.
(2) From the perspective of input data, the model dimensions of LMMs restrict the input matrix shape changing at large steps, \eg 4096 in Qwen-VL.
By doing so, the search space can be reduced up to ~20 times.
For example, the total search space of Qwen-VL on A100 is ~50,000 configurations, calculated by 288 configurations (A100's 36 common thread block shapes $\times$ 4 warp configurations $\times$ 2 instruction shapes) according to CUTLASS documentation~\cite{cutlass} and Qwen-VL model's 2048 maximum context length, reduced down to ~3,000.
Appx.~\ref{sec:SearchAlg_detail} shows the detailed algorithm.
With our algorithm, searching all optimal configurations for vision tasks with Qwen-VL on A100 GPU only costs <30 minutes.

\subsection{Flexible LoRA Adapters Orchestration}\label{sec:LoRAOrchestration}

To meet distinct performance requirements of vision applications, we propose an orchestrator to schedule requests, manage LoRA adapters, and switch inference modes, to enable efficient and flexible LoRA LLM inference runtime.
We first implement two tools with ATMM, a swift mode switcher and a mixture inference mode, to facilitate the orchestrator. 


\begin{figure}[t]
\centering
\includegraphics[width=0.35\textwidth]{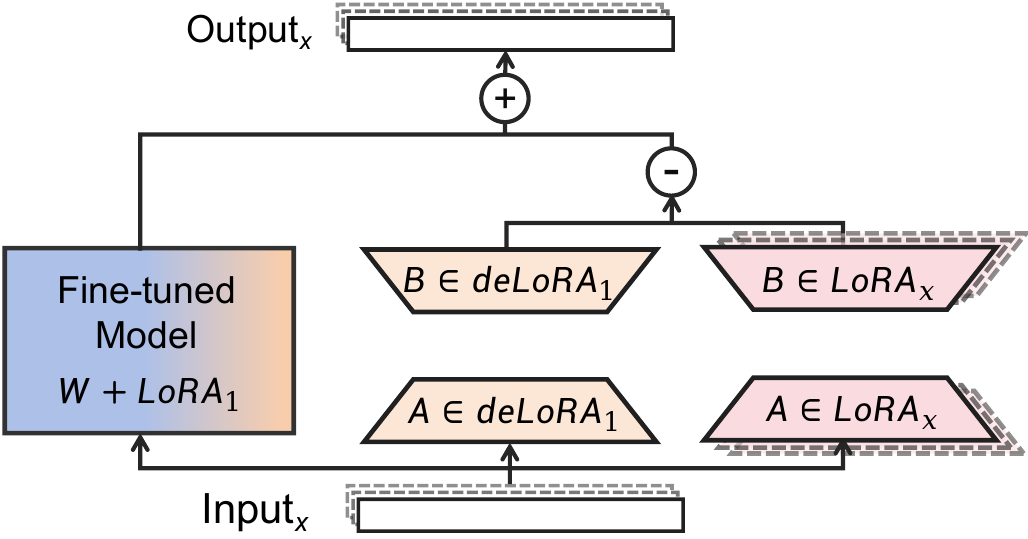}
    \caption{Mixture mode allows simultaneous execution of unmerged and merged modes to alleviate starvation.}
    \label{fig:deLoRA}
    \vspace{-1em}
\end{figure}

\subsubsection{Swift inference mode switch.}\label{sec:Switch}
As discussed in \S\ref{sec:challenges}, prior systems (\eg dLoRA~\cite{wu2024dlora}) introduce excessive extra latency during mode switch. 
\tb{To reduce these overheads, one common method is pre-computing LoRA matrices (\ie, matrix $\Delta W$, $B\times A$) for the entire base model and storing them in main memory, then swapping into GPU when needed.
However, such a mass of data, \eg $\sim$3GB per LoRA of QWen-VL-7B\footnote{Layer\#$\times\Delta W\times$precise, 32$\times$4096$\times$4096$\times$FP16, $\Delta W$'s shape is same as LMM.}, 
incurs very high delay, $\sim$1s each, on swapping.
Conversely, our swift mode switcher computes LoRA matrices at runtime and stores adapters, \ie $A$ and $B$, only 43MB each, on GPU. 
It has two core designs. 
(1) Eliminate unnecessary memory copy with contiguous memory allocation.
Our switcher pre-allocates contiguous memory for weight matrices,  
avoiding the memory copy in tensor reshape, enables efficient in-place LoRA matrices un-/merge.
(2) Compute all-layer LoRA matrices and un-/merge them in one shot.
With ATMM, the switcher can efficiently compute LoRA matrices of the entire model and add/subtract all of them onto/from the base mode weights in one shot.}
By doing so, our mode switch costs only <10ms, which speeds up dLoRA >5$\times$.



\subsubsection{Mixture inference mode.}\label{sec:deLoRA}
Compared to the unmerged mode, merged inference supports only one LoRA adapter at once, which results in the starvation of requests from other vision tasks.
To alleviate starvation, we propose a novel inference mode, deLoRA, to enable the simultaneous execution of merged and unmerged inference. 
As shown in Fig.\ref{fig:deLoRA}, LoRA$_1$ handles the requests in merged mode, while other LoRA adapters, LoRA$_x$, process their requests in unmerged mode. 
To maintain the consistent results of LoRA$_x$'s request, we introduce the deLoRA branch to prevent contamination from LoRA$_1$.
The weight of deLoRA is the same as that of the merged LoRA.
Based on the distributive property of matrix multiplication, the correctness can be verified as follows.
\begin{equation}\small
\setlength{\abovedisplayskip}{5pt}
\setlength{\belowdisplayskip}{5pt}
\begin{aligned}
\mathbf{output_x}&=\mathbf{input_x} \times (\mathbf{W_{merge}} - \mathbf{W_{deLoRA_1}} + \mathbf{W_{LoRA_x}}) \\ 
&=\mathbf{input_x} \times (\mathbf{W_{base}} + \mathbf{W_{LoRA_x}}), \notag
\end{aligned}
\end{equation}
in which $\mathbf{W_{merge}}=\mathbf{W_{base}} + \mathbf{W_{LoRA_1}}$,  $\mathbf{W_{LoRA_1}}=\mathbf{W_{deLoRA_1}}$, and $\mathbf{W}$ means the weight of the base model, deLoRA, and LoRA adapter; $\mathbf{input_x}$ and $\mathbf{output_x}$ are the input and output of the request of LoRA$_x$.
Overhead of computing $W_{deLoRA_1}$ is exactly the same as $LoRA_1$, and subtracting it introduces very few overhead.

Mixture inference mode shows two advantages.
1) It does not incur mode-switching costs from merged to unmerged mode.
2) It costs less extra computation than unmerged inference when there are more requests for the merged LoRA adapter than others.

\begin{algorithm}[t]
\footnotesize
\caption{Scheduling Policy}
\label{alg:scheduler}
\begin{algorithmic}[1]
\algnotext{EndIf}
\algnotext{EndFor}
\algnotext{EndFunction}
\Require Request $R\!=\!\!\left\{r_1,...,r_n\right\}$, LoRA adapter $L\!=\!\!\left\{l_1,...,l_n\right\}$, 
Infer mode $M$
\Ensure The batch of requests to be executed $B_{next}$
\Function{Scheduling}{R, M, L}  
\State $R_{starve} = \left[ R_i.credit>\theta \; for \; R_i \; in \; R \right]$ 
\State $len=MaxBS-|R_{starve}|$
\State $R_{merge}=argmax_{l_i\in L}\left\{ r_i\in R | r_i.lora==l_i \right\}$

\If{$|R_{starve}|/MaxBS \leq 0.5$ and $|R_{merge}|/MaxBS > 0.5$}
    \If{$|R_{starve}|==0$}
        \State $M=Merge$ and \Call{ModeSwitch}{$M$, $R_{merge}.lora$}
    \State $B_{next}=R_{merge}\left[ :MaxBS\right]$
    \Else
        \State $M=Mix$ and \Call{ModeSwitch}{$M$, $R_{merge}.lora$}
        \State $B_{next}=R_{starve} + (R_{merge}-R_{starve})\left[ :len\right]$
        \State \Call{InitDELoRA}{$B_{next}$}      
    \EndIf
\Else
    \State $M=Unmerge$ and \Call{ModeSwitch}{$M$, }
    \State $B_{next}=R_{starve} + (R-R_{starve})\left[ :len\right]$
\EndIf

\State \textbf{return} $B_{next}$
\EndFunction
\end{algorithmic}
\end{algorithm}
\setlength{\textfloatsep}{10pt}

\subsubsection{Scheduling policy.}
To minimize the average response latency and meet each request's latency constraint, our orchestrator must carefully orchestrate requests, adapters, and inference modes. 
Our policy follows a greedy heuristic 
which includes two principles.
(1) Executing in merged mode whenever possible, as it produces the fastest response and without extra overhead.
(2) When starvation occurs, switch to mixture first, then unmerge mode, in order of the switching cost and extra computation.
Alg.\ref{alg:scheduler} shows the pseudo-code. 
To alleviate starvation, it assigns each request a credit, indicating its waiting time adds the execution time in current mode and the mode switch latency (line \#2),  
and sets a tolerance threshold $\theta$ as the mixture mode condition. 
When the request workload meets the criteria for switching to merge mode, the algorithm switches the mode to merge (line \#5-8). 
When the number of starving requests exceeds $\theta$, Alg.\ref{alg:scheduler} processes them with mixture mode immediately (line \#9-12).
When it further exceeds half the maximum batch size, it switches to unmerge mode (line \#13-15).

One may think that a static inference mode by one-shot search can work well without switching.
However, it is hardly that in practice.
Most applications experience dynamic workloads, making it difficult to define an optimal execution mode or order.
For example, the video analytics application serves multiple users. 
The workload changes when new registered tasks (specified stream, tasks/adapter, accuracy) arrive for video analytics application, 
and the workload of LoRA adapters changes along with the multi-round VQA for visual retrieval application.
Detailed results can be found in \S\ref{sec:ComponentwiseAnalysis}.

\section{Implementation}\label{sec:Implementation}
We implement \name upon several tools, including Pytorch (v2.0.1) \cite{pytorch}, Triton (v2.0.1) \cite{Triton}, CUTLASS (v3.5.1)\cite{cutlass}, and vLLM (v0.3.0)~\cite{vLLM}, with $\sim$7.1K LOC. 
Most of the code is implemented in Python (v3.9) except the ATMM in CUDA.
We use vLLM, especially the LightLLM ~\cite{LightLLM} version, to build \name because of its advanced features, such as PagedAttention, iteration-level scheduling, and token-based memory management.
The three key techniques in \name
are implemented as follows.
(1) We implement the accuracy-aware LoRA adapter generation for popular LLMs, including Qwen-VL~\cite{QwenVL} and LLaVA~\cite{llava} series, based on transformers~\cite{transformers} and PEFT~\cite{peft} library.
(2) We implement ATMM using CUDA C++ based on CUTLASS. 
Its hash table, which stores optimal input-tiling pairs, is implemented with a 128-bit unsigned integer as a key to map the input shapes.
Since CUTLASS operators cannot be dynamically compiled, we created a Python interface to bind and package the code implementation of optimal tiling configurations with Pybind11~\cite{pybind11} and setuptools~\cite{setuptools}.
(3) We integrate ATMM into vLLM to support unmerged inference, mixture inference, and swift inference mode switch.
To manage the complicated adapters and requests in unmerged and mixture mode, we transform the LoRA type of each request into a one-hot vector and build a request-type mapping matrix of the current batch.
For efficient memory management, we use the unified memory management~\cite{sheng2023slora} for KV cache and adapters.
\name streams text and images in requests asynchronously via RPyC~\cite{rpyc}.
After receiving a request, \name identifies its LoRA adapter, dispatches it to the adapter, then generates a response and returns it.

\textbf{LoRA adapter swap.}
Considering it may generate an amount of LoRA adapters, we leverage the main memory.
To reduce delay, we only store $A$ and $B$, not $\Delta W$, and swap them asynchronously between the GPU and host memory, and compute $\Delta W$ with ATMM at runtime.

\textbf{Prefix caching.}
The same images may be accessed multiple times in some applications, \eg the multi-round visual question-answering~\cite{vqa2}. 
We implement a prefix caching, based on CacheBlend~\cite{yao2024cacheblend} and SGLang~\cite{zheng2023efficiently}, to reuse the same images' KV cache, avoiding redundant storage. 

\vspace{-1em}
\section{Evaluation}\label{sec:evalutaion}
We evaluate \name with two vision applications involving five tasks on three LMMs. 
The key takeaways are:

\scalebox{0.8}{$\bullet$} \name decreases 20-89\% end-to-end latency compared to state-of-the-art serving systems and achieves comparable accuracy with small models on specific domains. (\S\ref{sec:E2EPerf})

\scalebox{0.8}{$\bullet$} Accuracy-aware LoRA Adapter Generation brings remarkable accuracy and throughput benefits; Adaptive-tiling LoRA Adapters Batching and Flexible LoRA Adapters Orchestration boost great latency and throughput. (\S\ref{sec:ComponentwiseAnalysis})

\scalebox{0.8}{$\bullet$} \name shows strong stability to diverse workloads and LoRA adapter numbers, as well as scalability to multiple GPU resources. (\S\ref{sec:Scalability})

\begin{figure*}[t]
    \centering
    \begin{minipage}[b]{0.9\linewidth}
    \centering
    \includegraphics[width=0.9\linewidth]{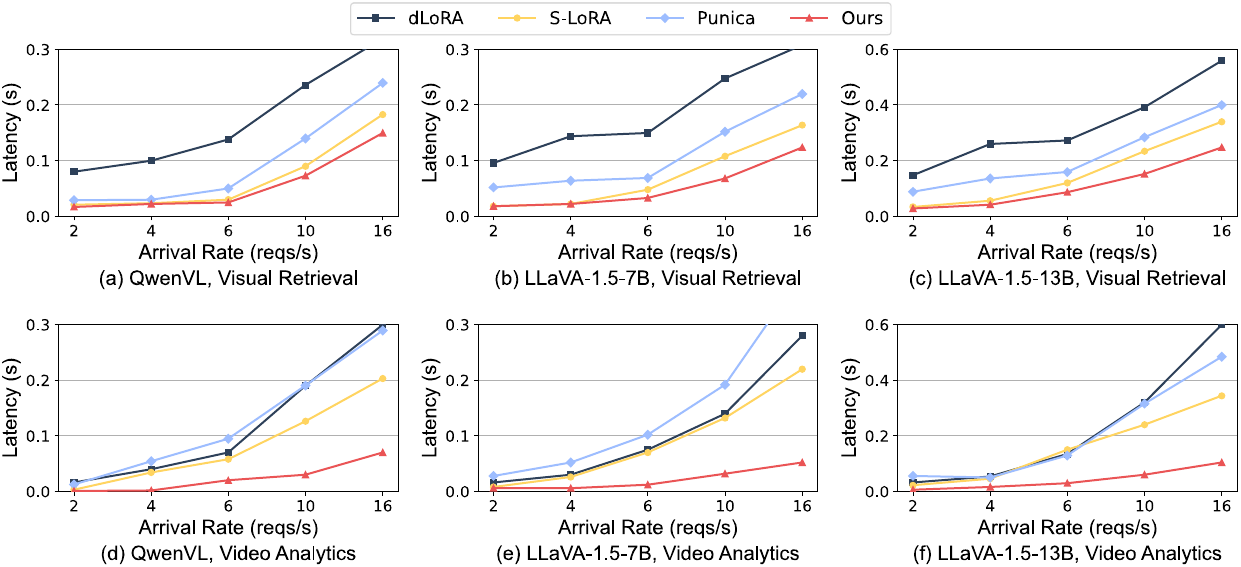}
    \end{minipage}
    \vspace{-1em}
    \caption{Average token latency comparison over various serving systems on two vision applications and three LMMs.}
    \label{fig:VQAcompare}
    \vspace{-1em}
\end{figure*}
\subsection{Experimental Setup}\label{sec:ExperimentalSetup}
\textbf{Vision applications and datasets.} 
We select two distinct types of visual applications: visual retrieval and video analytics, to evaluate the performance of \name.
Visual retrieval aims to analyze images and respond to queries. 
It involves visual question-answering, image caption, and specific-target detection tasks when needed by queries.
We evaluate visual retrieval on SharedGPT-4V~\cite{SharedGPT4V} and RefCOCO~\cite{kazemzadeh2014referitgame, yu2016modeling} datasets.
Video analytics ingests and analyzes each RGB frame from the video, then outputs results of fixed vision tasks, including object detection and video understanding like prior work~\cite{wang2024region, yuan2022PacketGame}.
Object detection locates and identifies objects on each video frame on YODA \cite{xiao2021towards} and Cityscapes \cite{Cordts2016Cityscapes}. 
Video understanding recognizes actions on consecutive video frames on UCF101~\cite{UCF101}. 

\noindent
\textbf{Testbed and workload.} 
We evaluated \name on a server equipped with one NVIDIA A100 80GB GPU and Intel Xeon Platinum 8358 CPU, with 128 GB of host memory.
The workload of visual retrieval is from production traces from the Microsoft Azure LLM inference trace 2023~\cite{AzureLLM}.
As the high volume of requests in this workload trace exceeds current hardware capabilities, like prior work~\cite{wu2024dlora}, we randomly sample requests at varying rates in a round-robin approach.
Like prior work~\cite{wang2024region,du2020server}, the video analytics workload ingests one video chunk per second, 30 frames each, each stream.
Our scope is primarily focused on optimizing single-GPU, single-instance LMM services.
If not specifically mentioned, the experiments are conducted on one instance of an A100 as reported above.

\noindent
\textbf{Metrics.} 
Accuracy metrics follow the standard metrics. 
Visual retrieval is measured by vqa-score \cite{goyal2017making}, object detection is evaluated by average F1-score and video understanding is evaluated via Top-1 accuracy. 
To the system metrics, we evaluate the end-to-end throughput, \ie requests per second, and the average token latency, \ie the sum of each request’s end-to-end latency divided by the total number of tokens.


\noindent
\textbf{Models.} 
We choose the widely-used open-sourced LMMs, Qwen-VL~\cite{QwenVL} and LLaVA-1.5 series~\cite{llava}.
We use various LLaVA models with different sizes, including LLaVA-v1.5-7B and LLaVA-v1.5-13B.
The details of these models are shown in Table \ref{tab:Models}. 
The rank of LoRA adapters is set to 64.
We also use five small models, VisionMamba~\cite{VisionMamba}, YOLO~\cite{glenn2021YOLOV5}, OSCAR~\cite{li2020oscar} VideoMAE~\cite{VideoMAE}, and UNINEXT~\cite{yan2023universal} that yield state-of-the-art performance for accuracy comparison on corresponding datasets.

\begin{figure}[t]
\centering
\includegraphics[width=0.45\textwidth]{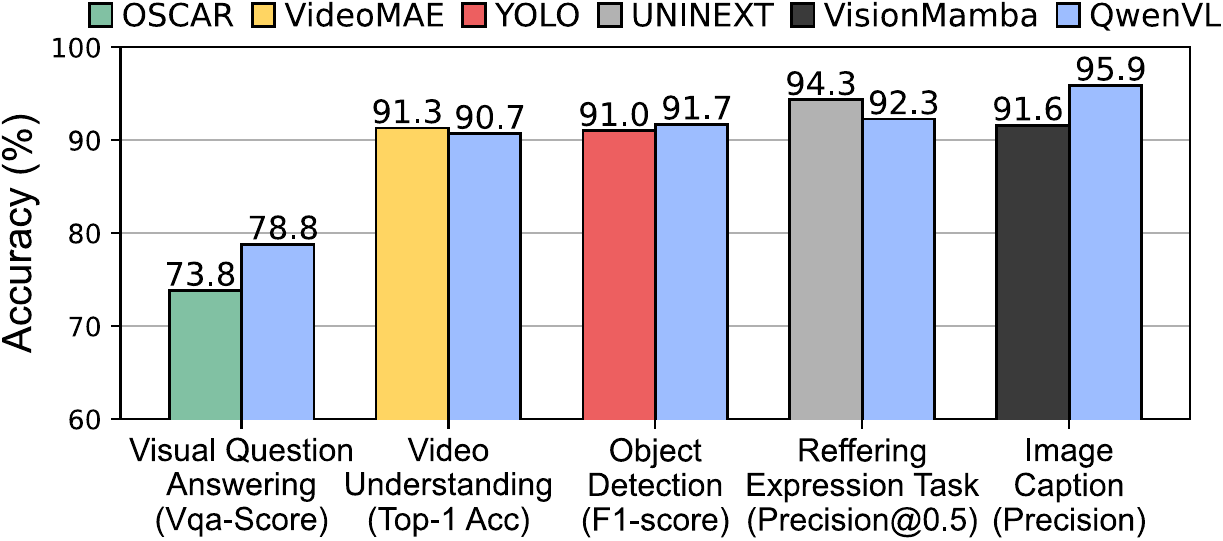}
\vspace{-1em}
    \caption{Accuracy comparison between small models and LMM model across five different vision tasks.}
    \label{fig:Accuracyexpriment}\vspace{-1em}
    
\end{figure}

\noindent
\textbf{Baselines.} To show the superiority brought by \name, we compare it with three different baselines:\\
\scalebox{0.8}{$\bullet$} 
\textbf{S-LoRA~\cite{sheng2023slora}} only serves in unmerge mode and employs its customized CUDA operator to batch concurrent heterogeneous LoRA computation. \\ 
\scalebox{0.8}{$\bullet$} 
\textbf{Punica~\cite{chen2024punica}} also serves in unmerge mode only, like S-LoRA, but employs its own operator.\\
\scalebox{0.8}{$\bullet$} 
\textbf{dLoRA~\cite{wu2024dlora}} dynamically switches between unmerged and merged mode based on workload and invokes PyTorch operator \texttt{Einsum} to implement unmerged inference.


\begin{table}[t] 
\footnotesize
\centering 
\begin{tabularx}{1\linewidth}{c c c c c}
\toprule 
\textbf{Model} & \textbf{Vision Encoder} & \textbf{Size} & \textbf{Layer \#} & \textbf{Dimension}  \\
\midrule 
\makecell{Qwen-VL-7B}  & \makecell{Openclip-ViT (1.9B)} & \makecell{18GB} & \makecell{32} & \makecell{4096}\\
\makecell{LLaVA1.5-7B}  & \makecell{CLIP-ViT (0.3B)} & \makecell{13GB} & \makecell{32} & \makecell{4096}\\
\makecell{LLaVA-1.5-13B}  & \makecell{CLIP-ViT (0.3B)} & \makecell{24GB} & \makecell{40} & \makecell{5120}\\
\bottomrule 
\end{tabularx}
\caption{Model configurations.} 
\label{tab:Models}
\vspace{-0.5em}
\end{table}

\subsection{End-to-End Performance}\label{sec:E2EPerf}
This section reports the E2E performance of \name on multiple LMMs and applications.

\begin{figure*}[t]
\centering

\hfill
\begin{minipage}[b]{0.23\linewidth}
\centering
\includegraphics[width=0.9\linewidth]{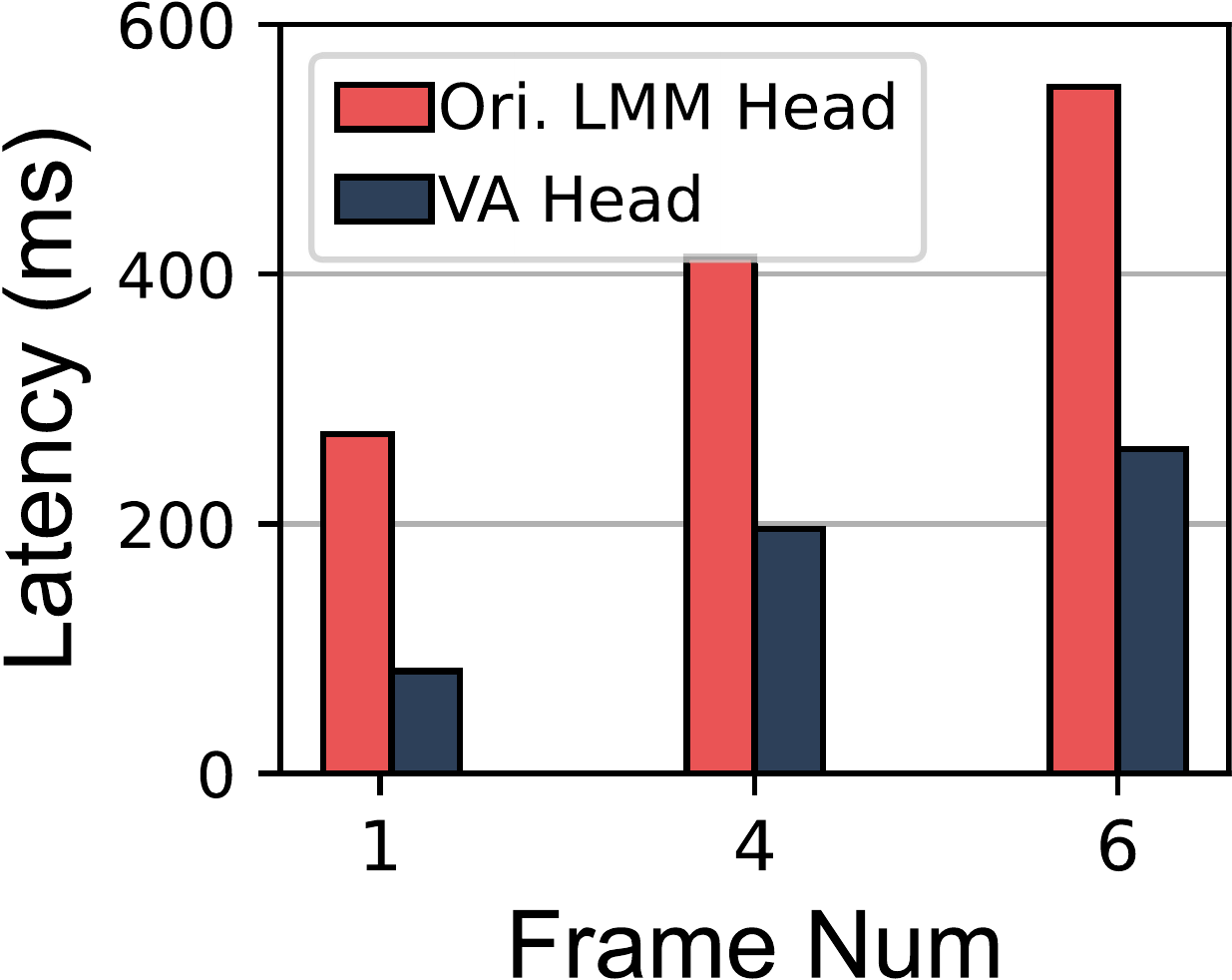}
\caption{Latency comparison between original LMM head and video analytics head.}
\label{fig:VAhead}
\end{minipage}
\hfill
\begin{minipage}[b]{0.235\linewidth}
\centering
\includegraphics[width=0.9\linewidth]{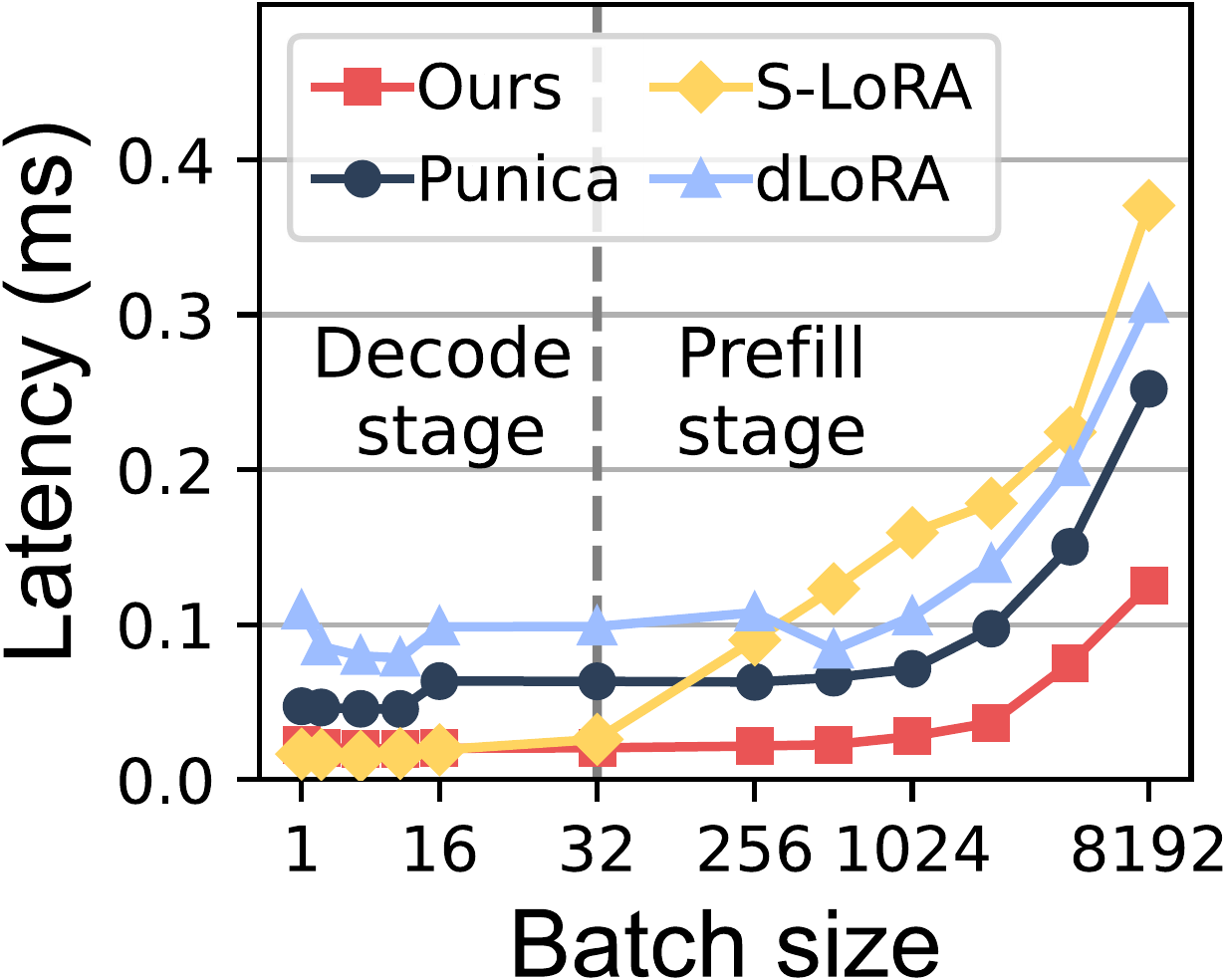}
\caption{Latency comparison of different operators across different token batch sizes.}
\label{fig:opcost}
\end{minipage}
\hfill
\begin{minipage}[b]{0.235\linewidth}
\centering
\includegraphics[width=0.9\linewidth]{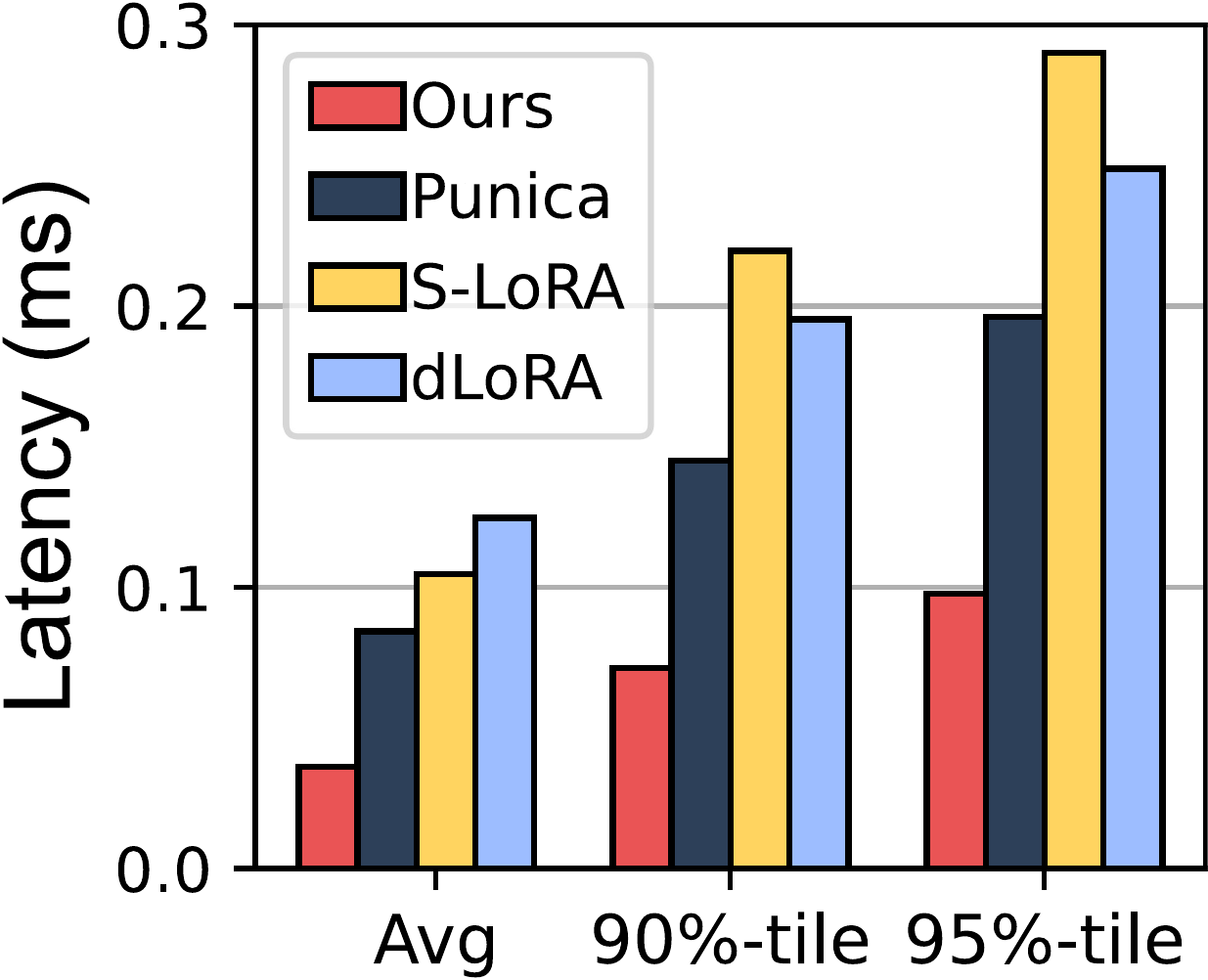}
\caption{\tb{Performance of different operators at average, 90-tile, and 95-tile .}}
 \label{fig:OperatorStat}  
\end{minipage}
\hfill
\begin{minipage}[b]{0.235\linewidth}
\centering
\includegraphics[width=0.9\linewidth]{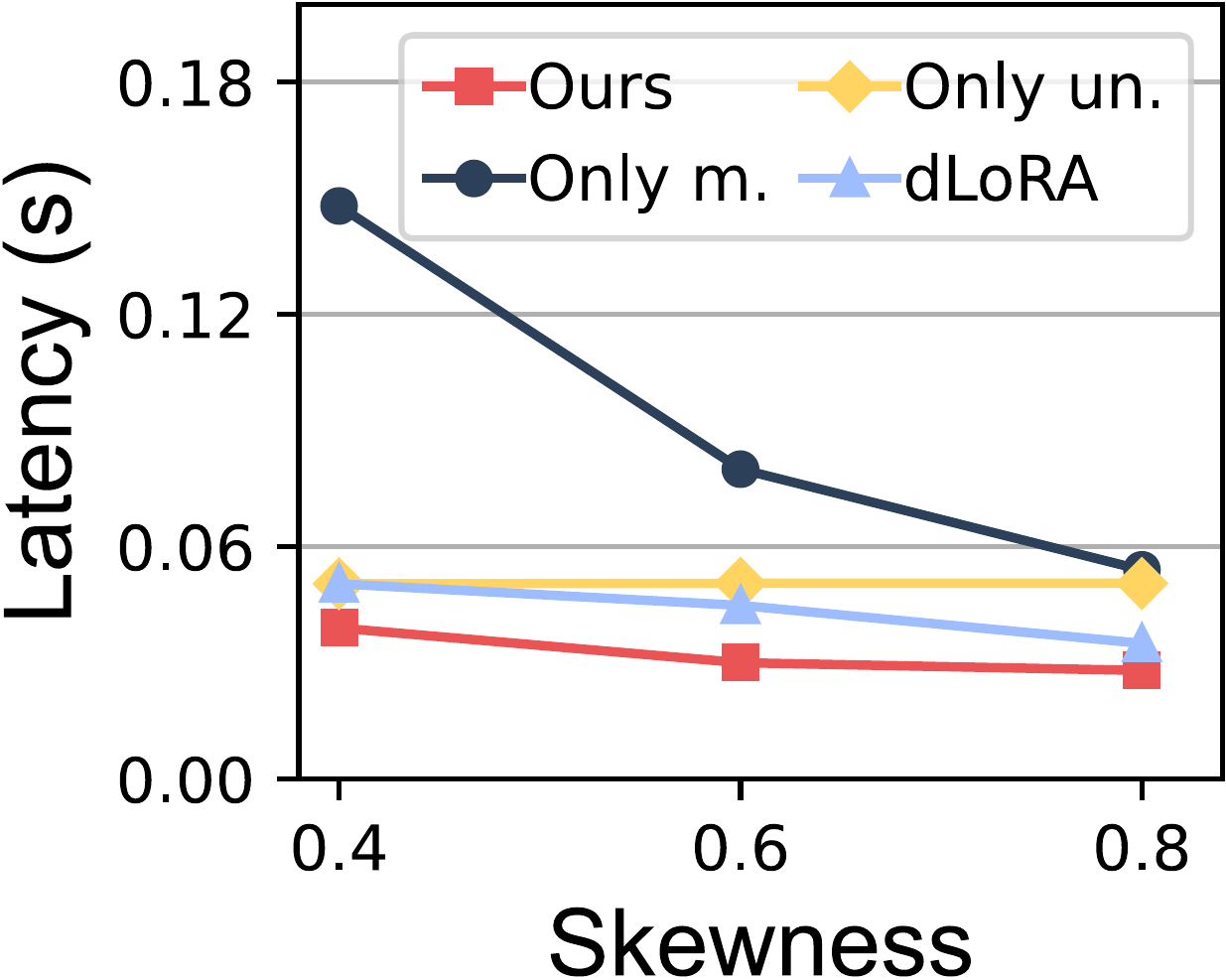}
\caption{\tb{Performance of different schedulers under different skewness.}}
\label{fig:scheduler}  
\end{minipage}
\vspace{-1em}
\end{figure*}

\noindent
\textbf{System performance.}
\name achieves notably lower average token latency than dLoRA, Punica, and SLoRA regardless of vision applications and LMMs. 
The first row in Fig. \ref{fig:VQAcompare} shows their performance for visual retrieval on three LMMs.
Across three LMMs, \name reduces 72\%, 50\%, and 20\% average token latency, compared to dLoRA, Punica, and S-LoRA, respectively.
To Punica and S-LoRA, this acceleration is obvious.
They only work in unmerge mode, which ignores the merge-friendly workload pattern, \eg 60\% of requests asking for the same LoRA adapter.
dLoRA, though, takes the preference of both inference modes into account, its high mode switching cost and inefficient \texttt{Einsum} operator in unmerged inference incurs this 20\% drop; 
conversely, \name's ATMM operator enables swift switch and efficient unmerge and mixture mode (more experiments in \S\ref{sec:ComponentwiseAnalysis}).
Moreover, along with the increasing requests per second, the inflection points of most serving systems occur at 6, which can be eliminated if equipped with more GPUs (more in \S\ref{sec:Scalability}).






\name delivers the best service on video analytics than other systems, as shown in the second row of Fig. \ref{fig:VQAcompare}. 
It makes 89\%, 83\%, and 71\% of average token latency reduction than dLoRA, Punica, and S-LoRA, respectively.
This benefit arises from the vision task head. 
It effectively eliminates the multi-rounds inference in the autoregressive manner of LLMs (more experiments in \S\ref{sec:exp_VisionTaskHead}).

Compare each column in Fig. \ref{fig:VQAcompare}.
\name produces more remarkable benefits than other serving systems on the video analytics application.
This difference stems from the different distribution of input and output token lengths in two vision applications.
Unlike visual retrieval, video analytics typically has fewer output tokens and more input tokens, \eg each video understanding request has 6$\times$256 input and 5-10 output tokens, while VQA has 256 and 200+.
Since input tokens can be batched computation in the prefill stage, they cost much less time (<1ms per token) than decode-stage output tokens (30-50ms per token). 
Furthermore, the extra latency of unmerged inference increases along with longer input length (See Fig. \ref{fig:opcost}); thus, longer-input video analytics can obtain more benefits from \name with mode switch.

\noindent
\textbf{Accuracy performance.}
\name achieves performance close to or surpassing the SOTA accuracy of domain-specific small models across various tasks. 
We report the results of Qwen-VL with fine-tuned LoRA adapters for different vision tasks as shown in Fig. \ref{fig:Accuracyexpriment}.
We conduct training on A100 80GB for five distinct tasks and test against corresponding SOTA small models. 
For instance, \name achieves a 4.3-5\% accuracy improvement in Visual QA and image captioning tasks. 
Additionally, for tasks where small models typically excel, such as object detection and video understanding, \name's fine-tuned LoRA adapters improve Qwen-VL 24.5-62.2\% accuracy, achieving competitive accuracy in these domains.

\subsection{Comprehensive Component-wise Analysis}\label{sec:ComponentwiseAnalysis}
We provide an in-depth performance analysis of individual system components.
If not mentioned, all results
are tested on the Qwen-VL model with 10 requests per second.

\subsubsection{Accuracy-aware LoRA Adapter Generation}\label{sec:exp_VisionTaskHead} helps \name achieve great throughput improvement while keeping high accuracy.
The accuracy gain has been discussed above; we only analyze the throughput gain from the vision task head.
Especially for video analytics tasks, the vision task head employed by \name significantly reduces the rounds of autoregressive decodes, thereby greatly enhancing system performance. 
As illustrated in Fig. \ref{fig:VAhead}, \name achieves a 41-63\% reduction in latency compared to the original language modeling head. 
This gain is attributed to the video analytics head's contribution to minimizing the prompt length and requiring only one inference round. 
In video understanding tasks, \name equipped with the video analytics head can match the accuracy of certain small models and handle 3-4 video streams in real time.

\subsubsection{Adaptive-tiling LoRA Adapters Batching} gives the most efficient and stable matrix multiplication by ATMM among all comparisons.
As shown in Fig. \ref{fig:opcost}, by testing over 100 rounds after 10 warm-ups on large amounts of diverse inputs, ATMM achieves the lowest average latency across different batch sizes, speeds up  2.7$\times$, 2.3$\times$, and 3.4$\times$ of S-LoRA, Punica, and dLoRA, respectively. 
On stability, the statistical results plotted in Fig. \ref{fig:OperatorStat} show that ATMM delivers the most robust performance, which \tr{reduce the latency fluctuation by 3$\times$, 2$\times$, and 2$\times$ compared to S-LoRA, Punica, and dLoRA. }
These benefits stem from the profile-based optimal tiling search at the offline phase. 
When the batch size exceeds 1024, for instance, ATMM adaptively adjusts to a larger tile shape, fully utilizing hardware resources, while other operators suffer from static tiling.

\begin{figure*}[t]
\centering
\setlength{\abovecaptionskip}{5pt}
\begin{minipage}[b]{0.24\linewidth}
\centering
\includegraphics[width=0.85\linewidth]{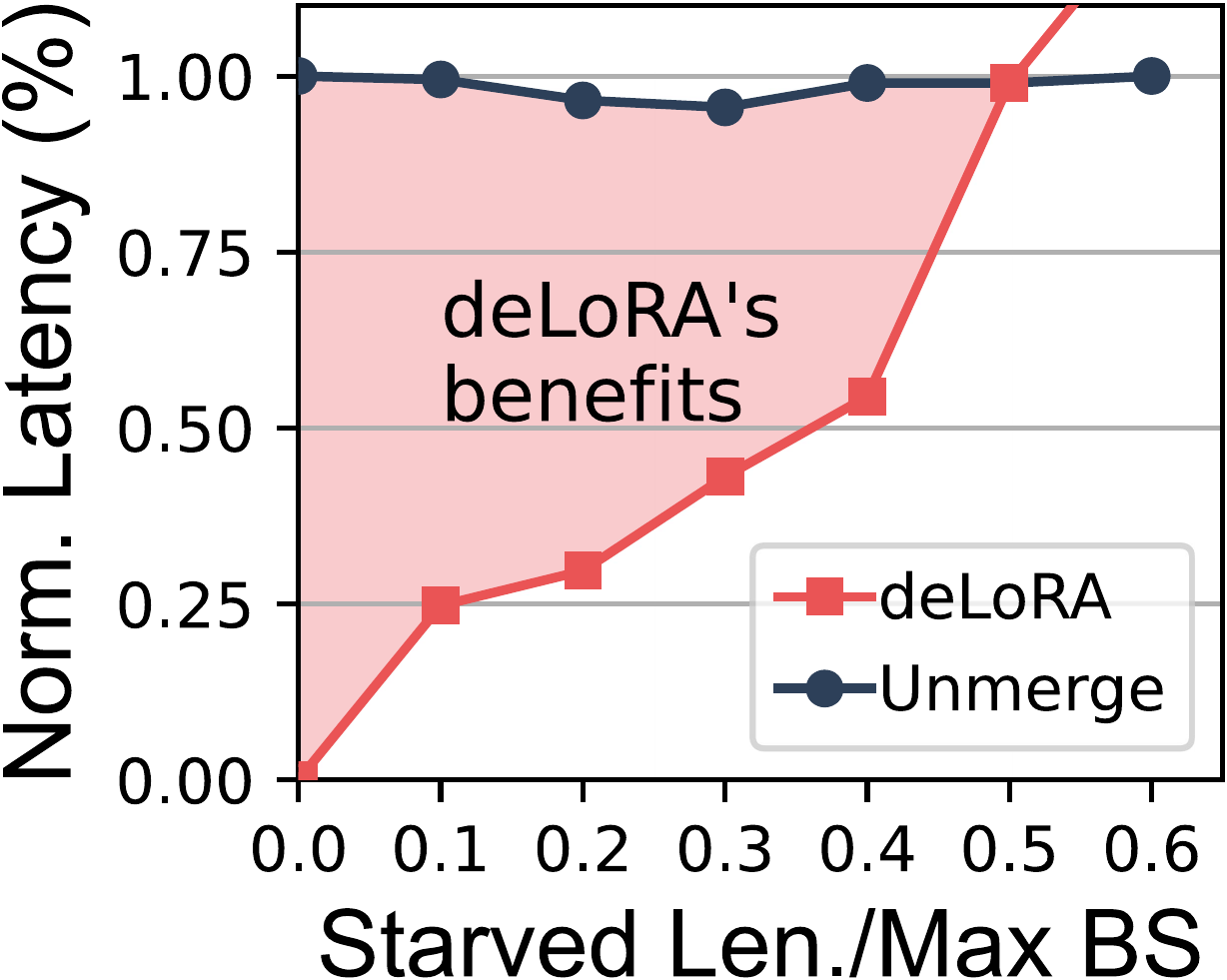}
\caption{Latency gain of mixture mode.}
\label{fig:delora}  
\end{minipage}
\hfill
\begin{minipage}[b]{0.24\linewidth}
    \centering
    \includegraphics[width=0.85\linewidth]{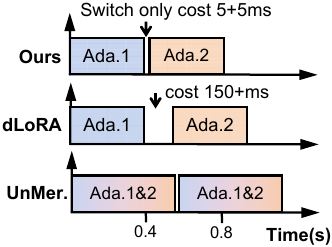}
    \caption{\tr{Benefits from swift mode switch.}}
    \label{fig:switch_exp}
\end{minipage}
  \begin{minipage}[b]{0.24\linewidth}
  \centering
  {\includegraphics[width=0.85\linewidth]{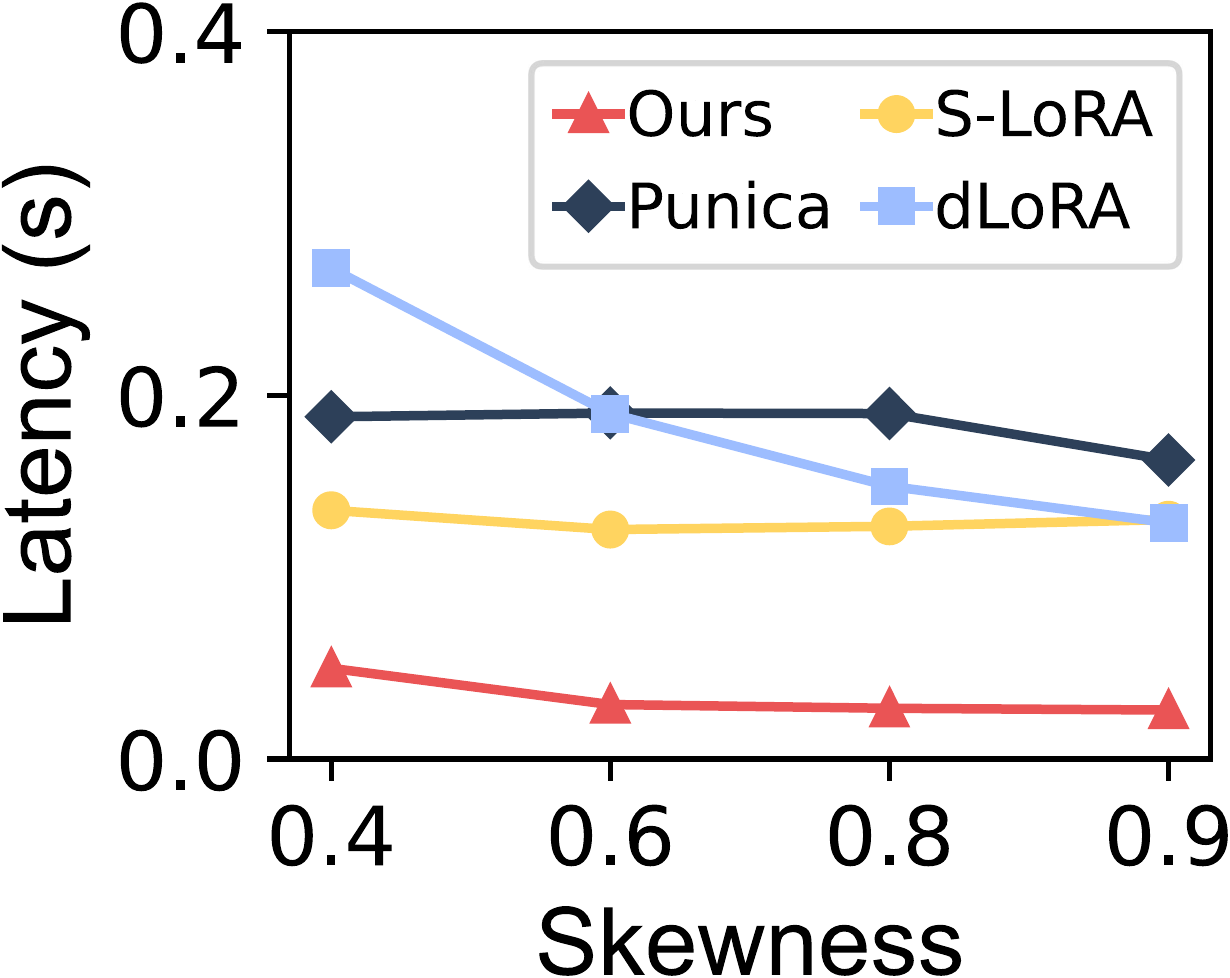}}
    \caption{Impact of the request skewness.
    }
    \label{fig:skewness}
    \end{minipage}
\hfill
  \begin{minipage}[b]{0.24\linewidth}
  \centering
    {\includegraphics[width=0.85\linewidth]{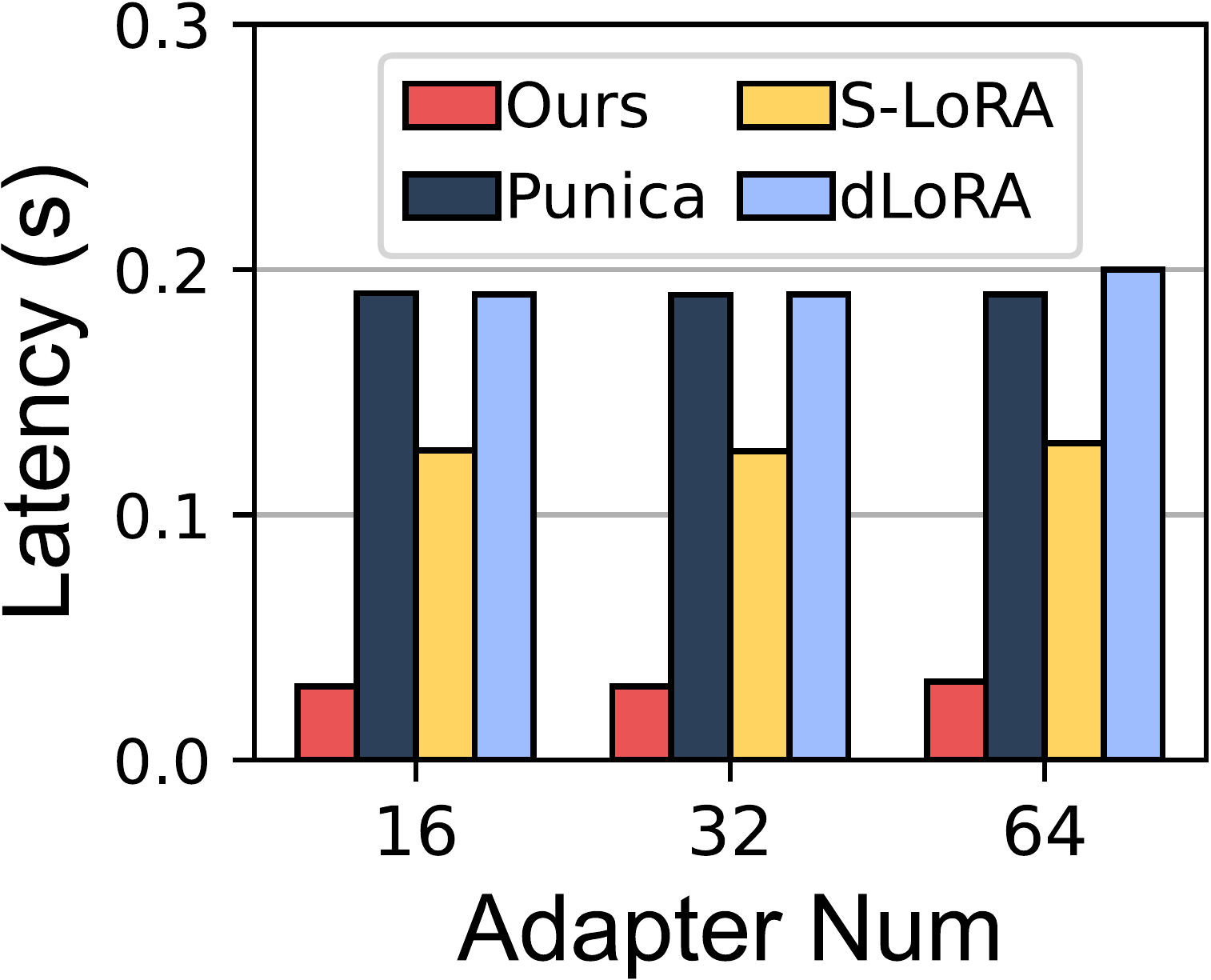}}
    \caption{Impact of the adapter number. 
    }
      \label{fig:adapternum}
  \end{minipage}
  \vspace{-1em}
\end{figure*}

At the decode stage with small input matrix shapes, the left part in Fig. \ref{fig:opcost}, ATMM maintains high efficiency by adapting smaller tile shapes. It delivers comparable latency to S-LoRA, outperforming dLoRA and Punica by 4.5$\times$ and 2.6$\times$.
dLoRA suffers from the large context-switching overhead of repeated kernel calls by \texttt{Einsum}, while Punica results in a low core utilization due to its mismatched tile shapes. 
Benefiting from the adaptive tiling, ATMM spends only 5ms to compute and un-/merge all-layer LoRA matrices.
\subsubsection{Flexible LoRA Adapters Orchestration} dynamically selects and switches inference modes with our swift switcher and deLoRA operation
offering the best service among comparisons. 
As shown in Fig. \ref{fig:scheduler}, \name outperforms merge only, unmerge only, and dLoRA by 33\%, 59\%, and 21\% of latency under different skewness, respectively.
The skewness indicates the proportion of the most required LoRA adapter.
Merge only processes requests invoking the same LoRA adapter, leading to underutilized resources and small batch sizes, while unmerge only introduces significant extra computation. 
dLoRA shows benefits only in highly skewed workloads as the poor performance of its unmerged inference operator \texttt{Einsum}.
Our scheduling policy performs the best because it fully utilizes low-latency merge mode, and invents mixture mode to eliminate some mode switch.

deLora significantly reduces the latency compared to unmerged inference as plotted in Fig. \ref{fig:delora}.
Its early execution for the starved requests 
saves an average of 62\% computation overhead when the starved requests' number is lower than 50\% of max batch size.
On the other hand, the swift inference mode switcher contributes a lot, too. 
Supported by ATMM, it yields 1.2$\times$ and 1.4$\times$ speed up compared to dLoRA and unmerge in Fig. \ref{fig:switch_exp} case that infers with two LoRA adapters.

\subsection{Stability and Scalability}\label{sec:Scalability} 
\name demonstrates great stability and scalability. 

\noindent
\textbf{Impacts of different skewness of requests.}
\name achieves the best average token latency compared to other systems under diverse skewness.
Fig. \ref{fig:skewness} shows that \name achieves a reduction in average token latency by 76-81\%, 72-83\%, and 63-76\% compared to dLoRA, Punica, and S-LoRA under four different skewness conditions. 
This superiority arises from the \name's timely mode switch and proper requests and adapters orchestration.
With the swift switcher and mixture mode, it responds to workload changes fast.

\noindent
\textbf{Impacts of different number of LoRA adapters.}
\name maintains the best and most stable performance when the number of LoRA adapters increases. 
As shown in Fig. \ref{fig:adapternum}, it suffers the minimal impact, which benefits from \name's efficient memory management. 
\name's pre-allocated contiguous memory reduces unnecessary memory copy or movement and memory fragmentation.
In addition, when the number increases to need LoRA to swap, \name's strategy swaps adapter and computes matrix at runtime, and the asynchronous swap contributes to the low latency.
With the highly optimized ATMM kernel, the LoRA matrix swapping 
keeps high stability compared to the matrix multiplication via batched GEMM in dLoRA.

\begin{figure}[t]
\centering
\setlength{\abovecaptionskip}{5pt}

  \begin{minipage}[b]{0.49\linewidth}
    \centering
    \footnotesize
    \begin{tabularx}{0.9\textwidth}{c | c}
      \toprule 
      \makecell{\textbf{GPU} \textbf{Num.}} & \makecell{\textbf{TPT} \textbf{(req/s)}} \\
      \midrule 
        1  &  6.07 \\
      \midrule 
      2  &  11.48\\
      \midrule 
      4  &  23.97\\
     \bottomrule 
    \end{tabularx}
    \vspace{15pt} 
    \captionof{table}{Scales to multiple GPUs.}
    \label{tab:MultiGPUTPT}
    \end{minipage}
\hfill
  \begin{minipage}[b]{0.45\linewidth}
  \centering
    {\includegraphics[width=0.9\linewidth]{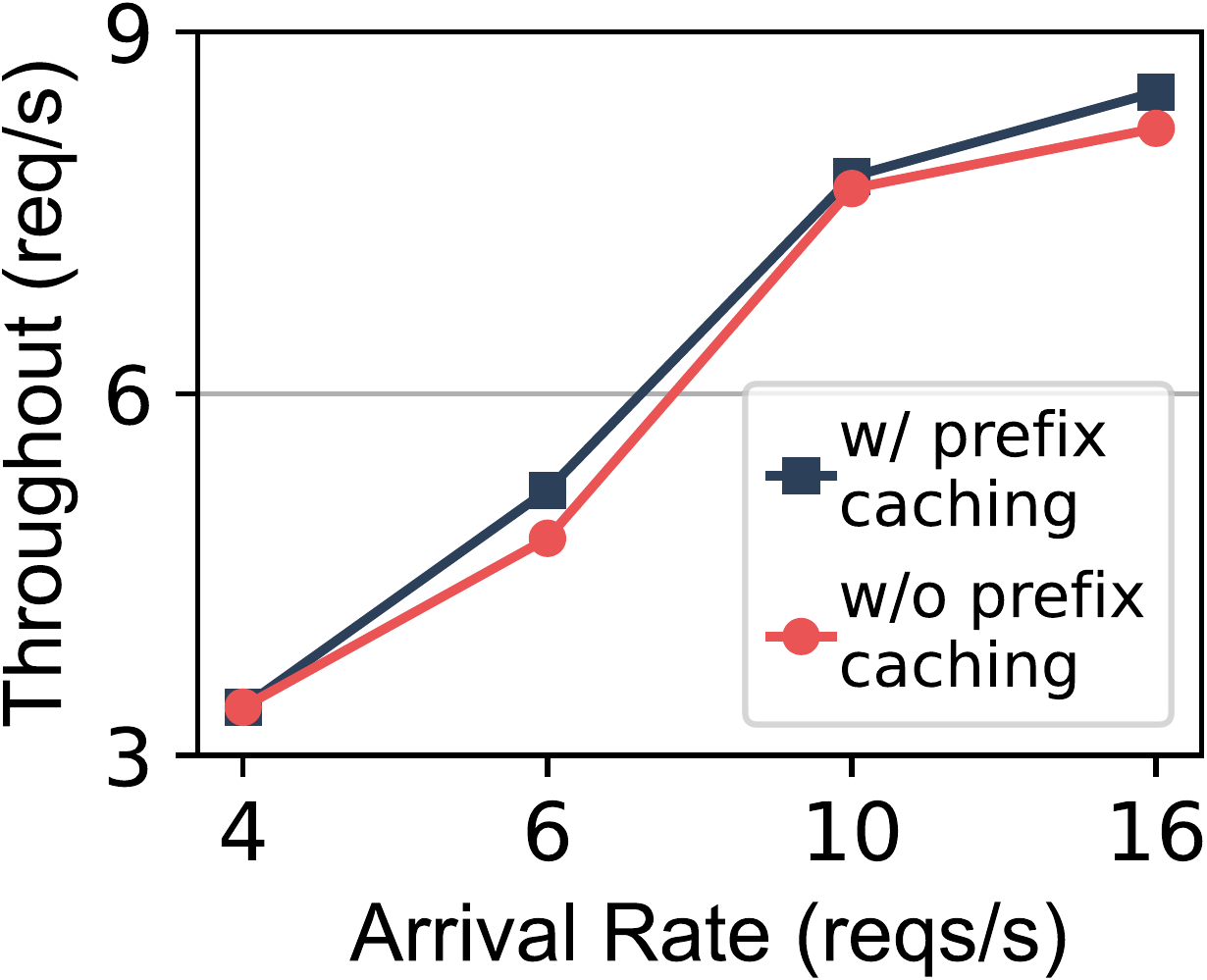}}
    \caption{Perf. breakdown of prefix caching.}
      \label{fig:prefixcachebreak}
  \end{minipage}
  \vspace{-1em}
\end{figure}

\noindent
\textbf{Scales to multiple GPUs.}
\name demonstrates excellent scalability and can significantly enhance the overall system throughput by multiple GPUs.
As shown in Tab. \ref{tab:MultiGPUTPT}, on servers equipped with 1, 2, and 4 A100 GPUs, the total system throughput can reach 6.07, 11.48, and 23.97 requests per second, respectively.
In future work, we can further improve system performance in multi-GPU scenarios by incorporating inter-GPU scheduling like dLoRA~\cite{wu2024dlora} and support larger LMM like InternVL2-76B~\cite{gao2024miniinternvlflexibletransferpocketmultimodal}. 

\noindent
\textbf{Impacts of prefix caching.}
VaLoRA maintains stable performance when removing prefix caching. 
As shown in Fig.~\ref{fig:prefixcachebreak}, VaLoRA loses less than 4\% of throughput after removing prefix caching. 
In VaLoRA, prefix caching is only a minor implementation supporting efficient multi-round VQA in visual retrieval applications.


    

\section{Disscussion and Limitations}\label{sec:Disscussion}
\textbf{Limited flexibility of vision task head.} 
The vision task head is deployed for one specific task and thus limits the flexibility.
For this reason, VaLoRA keeps the language modeling head for visual retrieval applications.
Even though, the vision task head, as a part of LoRA, facilitates efficient runtime by batching execution with ATMM.
We leave improving the flexibility to future work.

\noindent
\textbf{Knowledge fusion order.} 
The knowledge fusion order affects the final LoRA model quality, as illustrated in Fig.~\ref{fig:task_num}. 
Seeking the best order of training data is also an important but unsolved problem in the machine learning community, \eg curriculum learning and catastrophic forgetting problems in LLMs. 
In the future, we will borrow the idea from the ML community for knowledge fusion.

\noindent
\textbf{Knowledge pre-clustering.} 
Cluster knowledge before fine-tuning LoRA adapters can also affect the final quality.
VaLoRA clusters the data of the same vision task and keeps one LoRA adapter to serve one type of task.
We leave combining VaLoRA with other data clustering methods to future work.

\vspace{-1.2em}
\section{Related Works}\label{sec:RelatedWorks}
\textbf{Large model serving systems} recently leveraged system optimization techniques to improve LLM's inference efficiency. 
With paged thinking, vLLM~\cite{kwon2023efficient} proposes a PagedAttention operator cooperating with its block-based KV cache to minimize GPU memory fragmentation.
From advanced batching mechanisms, Orca~\cite{yu2022orca} introduces iteration-level scheduling to continuously batch requests of varying lengths, and DeepSpeed-FastGen~\cite{holmes2024deepspeed} further improves it with a dynamic split-fuse strategy. 
With a distributed architecture,
FlexGen~\cite{sheng2023flexgen} employs offloading to enhance LLM serving throughput, while 
Mooncack~\cite{qin2407mooncake} features a KVCache-centric architecture separating the prefill and decoding clusters.
With LLM inference characteristics, SpecInfer~\cite{miao2024specinfer} utilizes speculative decoding to reduce latency, while SARATHI~\cite{agrawal2023sarathi} schedules requests by piggybacking decodes and chunked prefills.

S-LoRA~\cite{sheng2023slora}, Punica~\cite{chen2024punica}, dLoRA~\cite{wu2024dlora} are the only three systems that also serve multiple LoRA LLMs by batching requests destined for different adapters. 
Their shortcomings are deeply analyzed in this paper. 
An earlier study, PetS~\cite{zhou2022pets}, also considers the scenario of serving multiple parameter-efficient DNN models, but it does not consider serving autoregressive LLMs and the unique system characteristics of LoRA adapters.
Compared to them, \name provides a more efficient serving runtime and, as an end-to-end system, includes LoRA adapter generation.

\noindent
\textbf{Parameter-efficient fine-tuning (PEFT)} \cite{hu2021lora, li2021prefix} 
is developed to adapt large pre-trained models to specific tasks or domains. 
By adjusting a few layers or parameters, PEFT retains core pre-trained knowledge while efficiently learning task-specific nuances~\cite{liu2022few}.
As a verification system, \name's LoRA adapter generation adopts a heuristic algorithm. 
It can be easily replaced by many advanced PEFT techniques~\cite{zhang2023adaptive}, and we leave this in future work.

\noindent
\textbf{Retrieval-augmented Generation (RAG)} \cite{borgeaud2022improving, lewis2020retrieval, jin2024ragcache} enhances LLMs by incorporating relevant knowledge from external databases, enabling comparable performance to fine-tuned LLMs~\cite{chen2024benchmarking}.
Some work \cite{ram2023context, trivedi2023interleaving} suggest
iterative retrieval throughout generation for higher quality.
\name's system performance greatly outperforms RAG because RAG's 
costly vector search \cite{chen2021spann} and long-context prompt. 

\vspace{-0.5em}
\section{Conclusion}\label{sec:Conclusion}

In this paper, we first explore the utilization of LMMs as foundation models for vision applications to achieve high serving efficiency and user-friendly nature language interface.
To achieve this, we propose \name, an end-to-end system that adapts LMMs for domain-specific visual tasks with LoRA adapters and efficiently manages them at runtime to enrich vision applications.
Across two typical vision applications, we show that \name enables the effective utilization of a single LMM to achieve superior performance and generalization in multiple visual tasks. 
While \name by no means is the final answer, we hope it serves as a stepping stone towards poly-basic design for future vision applications and demonstrates the potential of adapting LMM for visual tasks.

\section{Acknowledgment}\label{sec:Acknowledgment}

We would like to thank the anonymous reviewers and our shepherd Prof. Lai Fan, for their valuable guidance. 
Weijun Wang and Meng Li are the corresponding authors of this paper.
This work is supported by 
Carbon Neutrality and Energy System Transformation (CNEST) Program,  
NSFC (No.62402280, No.62272261, No.62272223), CPSF (No.20 24M761683), Shuimu Tsinghua Scholar Program (No.2023SM 201),
Tsinghua University (AIR)-AsiaInfo Technologies (China), Inc. Joint Research Center for 6G Network and Intelligent Computing, Wuxi Research Institute of Applied Technologies, Tsinghua University under Grant 20242001120, 
the New Generation Information Technology Innovation Project 2023 (2023IT196),
the Fundamental Research Funds for the Central Universities under Grant No. 2024300349,  
the Collaborative Innovation Center of Novel Software Technology and Industrialization, Nanjing University, 
the Jiangsu High-level Innovation and Entrepreneurship (Shuangchuang) Program,
and Beijing Academy of Artificial Intelligence (BAAI).



\bibliographystyle{plain}
\bibliography{reference.bib}

\appendix

\section{Profile-based Optimal Tiling Search Algorithm}\label{sec:SearchAlg_detail}

\begin{algorithm}\footnotesize
\caption{Tiling Search(model, hardware)}
\label{alg:InnerFree}
\begin{algorithmic}[1]
\algnotext{EndIf}
\algnotext{EndFor}
\algnotext{EndFunction}
\Require model, hardware
\Ensure Optimal configuration for each input shape
\Function{TilingSearch}{model, hardware}  
\State min, max=\Call{GetRange}{model, hardware} \Comment{\parbox[t]{.3\linewidth}{Limit the input size by limited memory}}
\State configlist=\Call{GetConfig}{harware} \Comment{\parbox[t]{.4\linewidth}{Limit the tiling step s.t. the arch characteristics}}\\
\For {shape \textbf{in} range(min, max, 32)} \Comment{\parbox[t]{.35\linewidth}{Reduce the search step size based on the vision task characteristics}}
    \For {rank \textbf{in} ranklist}
        \For {tbtile \textbf{in} configlist[0]}
            \For {warptile \textbf{in} configlist[1]}
                \State \Call{Profile}{shape, tbtile, warptile}
                \State \Call{UpdateBestConfig}{ }
            \EndFor
        \EndFor
    \EndFor
\EndFor
\EndFunction
\end{algorithmic}
\end{algorithm}

\section{Prompt Template of Vision Applications}\label{sec:PromptTemplate}
\begin{figure}[h]
    \centering
    \begin{minipage}[b]{1\linewidth}
    \centering
    \includegraphics[width=1\linewidth]{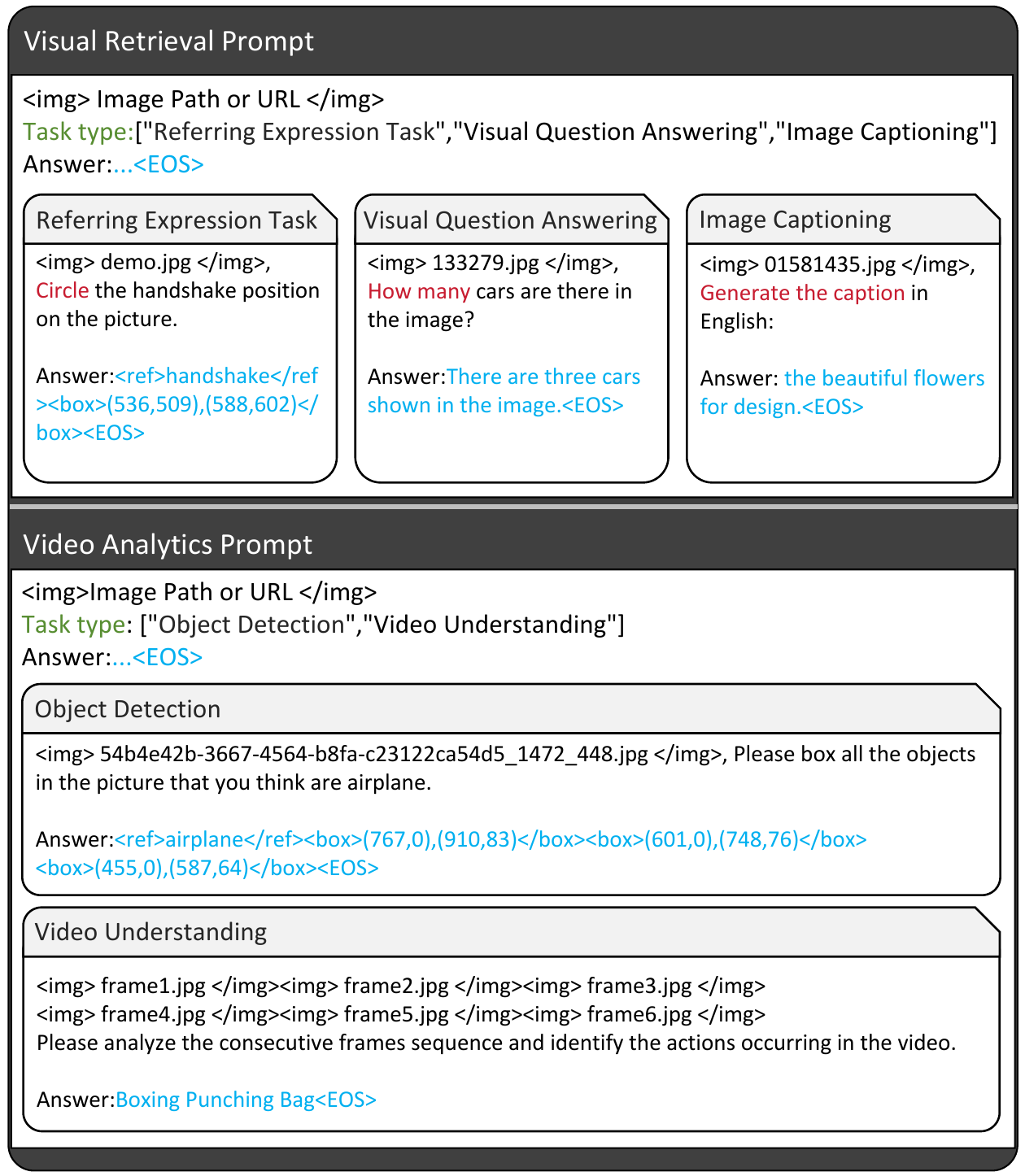}
    \end{minipage}
    \caption{The prompt template of vision applications.}
    \label{fig:templatepdf}
    \vspace{-1em}
\end{figure}
We report the prompt template for two vision applications, visual retrieval and video analytics, as shown in Fig. \ref{fig:templatepdf}. 
Visual retrieval includes three tasks: referring expression task, visual question answering, and image captioning. 
The black text represents the prompt, and the blue text shows the response. 
Video analytics involves two tasks: object detection and video understanding.
Object detection uses a similar prompt as the referring expression task.
Video understanding provides multiple image frames as input, followed by an instruction to analyze the actions depicted in the sequence.

\end{document}